%% file: main.tex
\documentclass[10pt,twocolumn,letterpaper]{article}

\usepackage[pagenumbers]{cvpr} %

\input{preamble}

\definecolor{cvprblue}{rgb}{0.21,0.49,0.74}
\usepackage[pagebackref,breaklinks,colorlinks,citecolor=cvprblue]{hyperref}

\usepackage[capitalize]{cleveref}
\crefname{section}{Sec.}{Secs.}
\Crefname{section}{Section}{Sections}
\Crefname{table}{Table}{Tables}
\crefname{table}{Tab.}{Tabs.}

\title{Odd-One-Out: Anomaly Detection by Comparing with Neighbors}

\author{Ankan Bhunia \quad Changjian Li \quad Hakan Bilen \vspace{2mm}\\
University of Edinburgh \vspace{1mm}\\
\href{https://github.com/VICO-UoE/OddOneOutAD}{https://github.com/VICO-UoE/OddOneOutAD}
}

\begin{document}

\twocolumn[{%
\renewcommand\twocolumn[1][]{#1}%
\maketitle
  \centering
  \vspace{-0.3cm}
  \includegraphics[width=\linewidth]{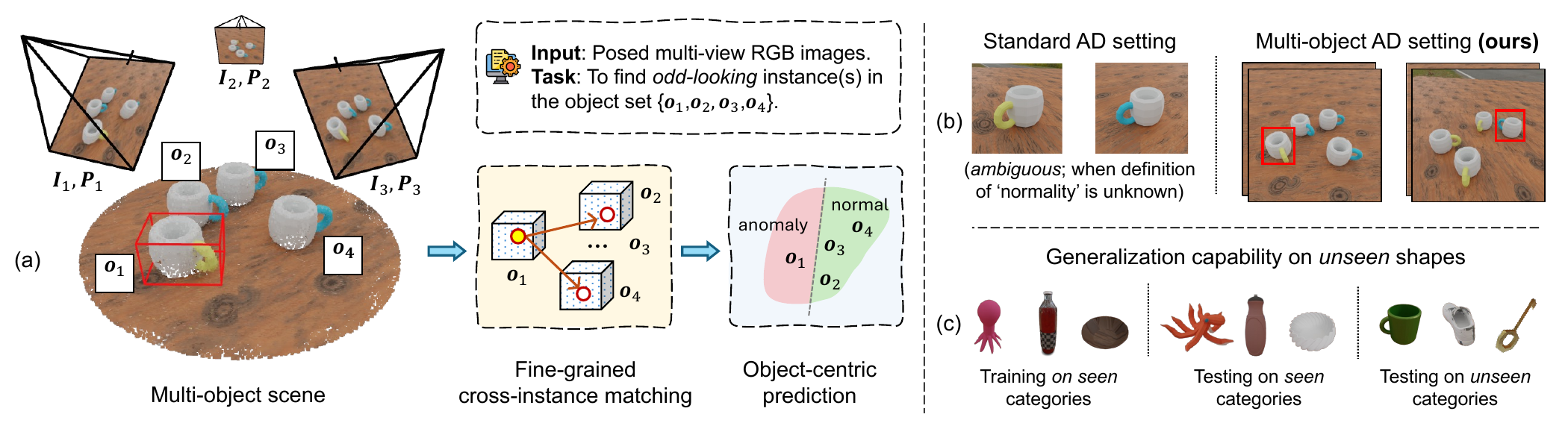}
  \captionsetup{type=figure}
  \vspace{-0.6cm}
  \captionof{figure}{(a) We propose a new anomaly detection task focused on identifying `odd-looking' objects relative to other instances within a scene. Inspired by real-world quality control in production environments, this task aims to detect subtle variations in geometry and texture, including defects like cracks and fractures, in a group of manufactured samples. (b) Our setting is scene-specific, requiring a comparison of object instances within the input scene, unlike the standard AD setting, which takes only a single object as input. (c) Our matching-based paradigm enables cross-category performance. %
  }
 \vspace{0.5cm}
  \label{fig:intro_big}
 }]

\input{sec/0-abstract}

\input{sec/1-intro}

\input{sec/2-related_work}

\input{sec/3-method}

\input{sec/4-experiment}

\noindent\textbf{Acknowledgment.} The authors acknowledge the support of Toyota Motor Europe.

{
    \small
    \bibliographystyle{ieeenat_fullname}
    \bibliography{main}
}

\input{sec/6-suppl}

\end{document}

%% file: preamble.tex
\usepackage[dvipsnames]{xcolor}

\usepackage[utf8]{inputenc} %
\usepackage[T1]{fontenc}    %
\usepackage{url}            %
\usepackage{booktabs}       %
\usepackage{amsfonts}       %
\usepackage{nicefrac}       %
\usepackage{microtype}      %
\usepackage{bm}
\usepackage{amsmath} 
\usepackage{graphicx}
\usepackage{bbm}
\usepackage{multirow}
\usepackage{booktabs}
\usepackage{subcaption}
\usepackage{booktabs}
\usepackage{colortbl}
\usepackage{booktabs} 
\usepackage{amssymb}%
\usepackage{pifont}%
\usepackage[accsupp]{axessibility}
\usepackage{wrapfig,lipsum,booktabs}
\usepackage{xspace}

\usepackage[accsupp]{axessibility}

\definecolor{mygray}{gray}{.9}

\makeatletter
\DeclareRobustCommand\onedot{\futurelet\@let@token\@onedot}
\def\@onedot{\ifx\@let@token.\else.\null\fi\xspace}
\def\eg{\emph{e.g}\onedot,\xspace} 
\def\ie{\emph{i.e}\onedot,\xspace}

\def \etal {{\emph{et al\onedot}\thinspace}}
\makeatother

%% file: sec/0-abstract.tex
\begin{abstract}
This paper introduces a novel anomaly detection (AD) problem aimed at identifying `odd-looking' objects within a scene by comparing them to other objects present.
Unlike traditional AD benchmarks with fixed anomaly criteria, our task detects anomalies specific to each scene by inferring a reference group of regular objects.
To address occlusions, we use multiple views of each scene as input, 
construct 3D object-centric models for each instance from 2D views, enhancing these models with geometrically consistent part-aware representations. 
Anomalous objects are then detected through cross-instance comparison.
We also introduce two new benchmarks, ToysAD-8K and PartsAD-15K as testbeds for future research in this task.
We provide a comprehensive analysis of our method quantitatively and qualitatively on these benchmarks.
  
\end{abstract}

%% file: sec/1-intro.tex
\section{Introduction}
Anomaly detection (AD)~\cite{chandola2009anomaly,pang2021deep} identifies patterns deviating from expected behavior.
In standard vision AD benchmarks, anomalies arise from high-level factors such as presence of a previously unseen category~\cite{chalapathy2018anomaly, ruff2018deep, ahmed2020detecting} or low-level ones such as defects and variations in the normal object shape and texture~\cite{carrera2016defect,bergmann2021mvtec,deecke2021transfer}.
These benchmarks typically assess objects in isolation, disregarding inter-object relationships  and contextual dependencies.
However, in many real-world scenarios, `normality' is often context-dependent.
For instance, in a production line for blue-handled coffee cups, a yellow handle is anomalous, and in a production line for yellow-handled coffee cups, a blue handle is anomalous (see \cref{fig:intro_big}(b)).
Methods assuming a fixed standard of normality cannot accommodate such context-dependent anomalies.

Inspired by real-world visual inspection, where efficiency in automatically processing large volumes of the same product is crucial, we introduce a new AD problem with along two new benchmarks and a novel solution.
Our setup (see \cref{fig:intro_big}(a)) consists of scenes containing multiple instances of the same object (\eg coffee cup) including either all normal, or a mix of normal and anomalous instances.
Each scene has at least two normal instances, which together serve as a scene-specific reference for `normality'.
The objective is to detect anomalous instances while generalizing to previously unseen scenes including unseen objects and layouts.
{This approach facilitates cost-effective models that do not require retraining for new products.}
Our task diverges from traditional AD tasks in two key ways:
(1) anomaly detection is scene-specific, requiring a comparative analysis of instances rather than solely relying on individual inspection -- only certain anomalies (\eg cracks and misaligned parts) are detectable at instance-level; (2) it involves multi-view scene analysis, capturing a comprehensive object model while addressing self-occlusion and inter-instance occlusion, unlike standard AD benchmarks that typically rely on single-view inputs (see \cref{fig:intro_big}(a)).

The proposed problem presents several unique challenges requiring the following key capabilities: 
i) \textit{3D-aware modeling:} constructing accurate representations of object instances by capturing their geometry and appearance from multiple viewpoints while handling occlusions,
ii) \textit{pose-agnostic comparison:} aligning and comparing object instances effectively  without their pose information during training and testing,
iii) \textit{generalizable representations:} learning representations that generalize to unseen object instances from both seen or unseen categories at test time.

To tackle these challenges, we introduce a novel method that integrates multi-view scene analysis by projecting them into a 3D voxel grid and generating 3D object-centric representations.
Our model classifies instances through cross-instance correlation using an efficient attention mechanism.
Leveraging recent advances in differentiable rendering~\cite{mildenhall2021nerf} and self-supervised learning~\cite{oquab2023dinov2}, we ensure robust 3D representation learning by rendering voxel representations for multiple viewpoints and aligning them with corresponding 2D views for geometric consistency.
Additionally, we enhance part-aware 3D representation learning by distilling features from 2D self-supervised model DINOv2~\cite{oquab2023dinov2}, improving correspondence matching and pose-invariant alignment.
As no existing benchmarks address this task, we introduce two new benchmarks: \emph{ToysAD-8K} including common objects categories, and \emph{PartsAD-15K} focusing on mechanical parts, to foster further research.
Our method significantly surpasses baselines that lack cross-instance comparison and part-awareness, and we provide an in-depth analysis of our model.

%% file: sec/2-related_work.tex
\section{Related Work}
\noindent \textbf{AD benchmarks.} A major challenge in AD research is the limited availability of large datasets with realistic anomalies. 
Earlier works~\cite{chalapathy2018anomaly, ruff2018deep} focusing on high-level semantic anomalies often use existing classification datasets by treating a subset of classes as anomalies and the remainder as normal. 
There are also datasets containing real-world anomaly instances; for instance, MVTec-AD~\cite{bergmann2019mvtec} includes industrial objects with various defects like scratches, dents, and contaminations, while Carrera~\etal\cite{carrera2016defect} presents various defects in nanofibrous material.
The VisA~\cite{zou2022spot} comprises complex industrial objects such as PCBs, as well as simpler objects like capsules and cashews, spanning a total of 12 categories. 
Zhou~\etal\cite{zhou2024pad} introduced a pose-agnostic framework by introducing the PAD dataset, which comprises images of 20 LEGO bricks of animal toys from diverse viewpoints/poses. 
Bhunia~\etal\cite{bhunia2024looking} proposed a conditional AD detection technique by comparing a given 3D reference shape to a single object in a query image.
In contrast, we target AD within multi-object, multi-view scene environments, where anomalies are defined relative to scenes and detected through cross-instance in the scene. 
Moreover, our method does not require 3D reference models which can be  costly and tedious to obtain.

\noindent \textbf{Few-shot AD.} 
There are some works~\cite{ding2022catching, huang2022registration, xie2023pushing, wu2021learning} aiming to detect anomalies from a small number of normal samples as support images. 
In our setting, the concept of normality in each image is also learned from a few instances only.
However, unlike them, the concept of normality is scene-specific, and our method can generalize to previously unseen instances without requiring any modification by learning from a support set. 
In addition, our setting involves multi-object multi-view data samples as input, unlike the single-object single-view in theirs.

\noindent \textbf{Multi-view 3D vision.} Multi-view 3D detection~\cite{rukhovich2022imvoxelnet, shi2021geometry, xie2022m, wang2022detr3d} is a related problem that aims to predict the locations and classes of objects in 3D space, given multi-view images of a scene along with their corresponding camera poses as input. 
Most existing works first project 2D image features onto a 3D voxel grid, followed by a detection head~\cite{lang2019pointpillars, yin2021center, carion2020end} that outputs the final 3D bounding boxes and class labels. 
While our task can be naively solved by treating it as a 3D object detection problem with anomaly and normal as the two possible classes, this approach %
has limited ability to perform effective comparisons with other instances in the scene, which is required for fine-grained AD. 
We compare our method to a 3D object detection technique in \cref{sec:exp}.
Another related area involves taking multi-view images as input and training a feedforward model for 3D volume-based reconstruction~\cite{murez2020atlas, yuan20233d, sun2021neuralrecon} and novel view synthesis~\cite{trevithick2020grf, wang2021ibrnet, yu2021pixelnerf, chen2021mvsnerf, jiang2022few}. Similarly, in this work, we focus on learning a feedforward model in a multi-object scene environment using sparse multi-view images but for AD.

\begin{figure*}[!ht]
\begin{center}
   \includegraphics[width=0.87\linewidth]{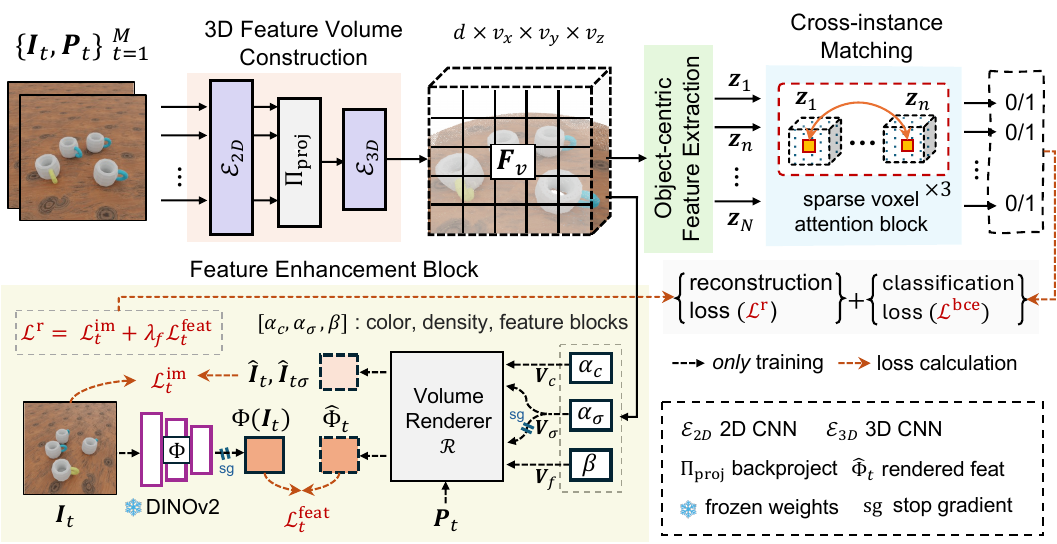}
\end{center}
\vspace{-4mm}
\caption{Overview of our framework. We extract features from a sequence of input views using a 2D CNN and back-project them into a 3D volume, which is then refined with a 3D CNN, resulting in $\bm{F}_v$. Next, we extract object-centric feature volumes $\{\bm{z}_n\}_{n=1}^{N}$, which are fed into the cross-instance matching module to learn correlations among objects using sparse voxel attention. To improve the 3D representation of the scene, we distill the knowledge of a 2D vision model namely DINOv2, and integrate the learned knowledge into our 3D network via differentiable rendering. This aids in obtaining a part-aware and geometrically consistent 3D feature representation. 
}
\label{fig:method}
 \vspace{-4mm}
\end{figure*}

\noindent \textbf{Leveraging large-scale models.} Large-scale pretraining on image datasets has shown impressive generalization capabilities on various tasks~\cite{radford2021learning, kirillov2023segment, oquab2023dinov2}. Previous works~\cite{zhang2024tale, banani2024probing} demonstrate that
features extracted from DINOv2~\cite{oquab2023dinov2} serve as effective dense visual descriptors with localized semantic information for dense correspondence estimation tasks. Some recent works~\cite{kobayashi2022decomposing, tschernezki2022neural} use feature distillation techniques to leverage 2D foundation vision models for 3D tasks. Inspired by these works, we utilize DINOv2 to distill its dense semantic knowledge into our 3D network enabling our network to infer robust local correspondences, which aids in fine-grained object matching.

%% file: sec/3-method.tex
\section{Method}

Consider a scene containing $N$ instances $\{o_{n}\}_{n=1}^{N}$ of the same rigid object where their poses are unknown. 
The goal is to identify anomalies in the group of objects by assessing their mutual similarity. 
An observation in our setting consists of $M$-view $H \times W$ dimensional RGB images $\mathcal{I}=\{\bm{I}_{t}\}_{t=1}^{M}$, and their corresponding camera projection matrices $\mathcal{P}=\{\bm{P}_{t}\}_{t=1}^{M}$ where $\bm{P}_{t}=\bm{K}\left[\bm{R}_{t}|\bm{T}_{t}\right]$, with intrinsic $\bm{K_t}$, rotation $\bm{R}_{t}$ and translation $\bm{T}_{t}$ matrices. We set $M$ to $5$, forming sparse-view inputs in our experiments. 
Our goal is to learn a mapping $\psi$ from the multi-view images to object-centric anomaly labels ${y}_{n}\in \{0,1\}$ and its corresponding 3D bounding box $\bm{b}_{n}$, defined as:
\begin{equation}
    \psi : \{(\bm{I}_{t}, \bm{P}_{t})\}_{t=1}^{M} \mapsto \{({y}_{n}, \bm{b}_{n})\}_{n=1}^{N}.
\end{equation}
Importantly $y_n$ is defined relative to other objects in the scene. 
For example, consider a group of three coffee cups, where two have {yellow} handles, and one has a {blue} handle. Here, the blue-handled cup is classified as an anomaly.
Notably, we do not assume that normal instances outnumber anomalies; rather, we assume that anomalies exhibit instance-specific variations (\eg, fractures differing in location and lengths), while normal instances maintain high visual similarity.

Our pipeline, illustrated in \cref{fig:method}, is comprised of four main components: First, the \textit{3D feature volume construction} module (\cref{sec:feature_contr}) encodes each view image and projects it to 3D, forming a fused 3D feature volume. 
Second, the \textit{feature enhancement} block (Sec.~\ref{sec:enhance}), employed \textit{only} during training, facilitates the learning of a part-aware and geometrically consistent 3D space through differentiable rendering and feature distillation.
Third, the \textit{object-centric feature extraction} module (\cref{sec:object_centric}) locates each object in the scene and represents it with a feature volume.
Finally, the \textit{cross-instance matching} module (\cref{sec:cross_matching}) efficiently compares all similar object regions using a sparse voxel attention mechanism and predicts instance anomalous labels and 3D coordinates. 
Next, we elaborate on details.

\subsection{3D Feature Volume Construction}
\label{sec:feature_contr}
In our first module, we obtain a 3D feature volume representation given multi-view images as input.
Specifically, we first extract 2D features $\bm{F}_{t} = \mathcal{E}_{2D}(\bm{I}_{t}) \in \mathbb{R}^{d \times h \times w}$ for each input view using a shared CNN encoder $\mathcal{E}_{2D}$, where $d$ is the feature dimension. These 2D features are then projected into 3D voxel space as follows:
\begin{equation}
   \bm{F}_{v} = \mathcal{E}_{3D}(\texttt{aggr}(\{\Pi_{\texttt{proj}}(\bm{F}_{t}, \bm{P}_{t})\}_{t=1}^{M})),
\end{equation}
where $\Pi_{\texttt{proj}}$ back-projects each view feature $\bm{F}_{t}$ using known camera intrinsic and extrinsic parameters, generating 3D feature volumes of size $d \times v_x \times v_y \times v_z$. %
These feature volumes are aggregated over all input views using an average operation as in~\cite{murez2020atlas, sun2021neuralrecon}. Finally, a 3D CNN-based network $\mathcal{E}_{3D}$ is employed to refine the aggregated feature volume, resulting in a final voxel representation $\bm{F}_{v}$ of size $d \times v_x \times v_y \times v_z$, which is further fed into the feature enhancement block, described next, to obtain a part-aware and geometrically consistent 3D representation.

\subsection{Feature Enhancement Block}
\label{sec:enhance}
We use volume rendering~\cite{mildenhall2021nerf} to reconstruct the geometry and appearance of the scene. 
We implement the rendering operation as in~\cite{jiang2022few}, where for each 3D query point on a ray, we retrieve its corresponding features by bilinearly interpolating between the neighboring voxel grids. 
Specifically, we first apply a two-layered $1\times1\times1$ convolution block, \ie $\alpha_c$ and $\alpha_\sigma$ respectively, to obtain color and density volumes denoted as ($\bm{V}_c$, $\bm{V}_\sigma$). 
Then, the pixel-wise color and density maps are composed by integrating along a camera ray using volume renderer $\mathcal{R}$. 
Following this, we compute the L2 image reconstruction loss $\mathcal{L}^\text{im}_t$. 
Formally, the image rendering and its corresponding loss function for a single viewpoint $\bm{P}_t$ are shown below:
\begin{equation}
\begin{split}
    \label{eq:renderer}
    \mathcal{L}^\text{im}_t = ||\bm{I}_t-\hat{\bm{I}}_t||^{2} + ||\bm{I}_{t\sigma}-\hat{\bm{I}}_{t\sigma}||^{2},
     \end{split}
\end{equation}
where $[\hat{\bm{I}}_t, \hat{\bm{I}}_{t\sigma}] = \mathcal{R}([\bm{V}_c, \bm{V}_\sigma], \bm{P}_t)$, $\hat{\bm{I}}_t$ and $\hat{\bm{I}}_{t\sigma}$ are the rendered image and mask. %

Moreover, we improve the 3D voxel representation $\bm{F}_{v}$ by reconstructing neural features in addition to geometry and appearance. %
We supervise the feature reconstruction by a pretrained 2D image encoder $\Phi$ as a teacher network. 
We choose DINOv2~\cite{oquab2023dinov2} as the teacher network due to its excellent ability to capture various object geometries %
and correspondences. 
To this end, we use a projector function $\beta$, instantiated as a four-layered $1\times1\times1$ convolution block that projects $\bm{F}_{v}$ to a \textit{neural feature field} $\bm{V}_f$, changing the channel dimension from $d$ to $d_f$. 
Similar to color rendering, taking  $\bm{V}_f$ and $\bm{V}_\sigma$ as input, we generate rendered features $\hat{\Phi}_t$ of size $d_f \times h_f \times w_f$ at a given viewpoint using volume renderer $\mathcal{R}$%
, and minimize the difference between the rendered features and the teacher's features $\Phi(\bm{I}_t)$. 
We choose cosine distance as our feature loss ($\mathcal{L}^\text{feat}_t$), which we find easier to optimize compared to the standard L2 loss. 
We do not allow the gradients to follow through the density in the rendering of features $\hat{\Phi}_t$, as the teacher’s features are not fully multi-view consistent~\cite{banani2024probing}, which could harm the quality of reconstructed geometry. 
The final reconstruction loss is the sum of all image and feature reconstruction losses: %
\begin{equation}
    \mathcal{L}^{r} = \sum_{t=1}^{M}(\mathcal{L}^\text{im}_t + \mathcal{L}^\text{feat}_t).
\end{equation}

The key benefits of reconstructing DINOv2 features are twofold. First, distilling features from general-purpose feature extractors pretrained on large external datasets incorporates open-world knowledge into the 3D representation. 
This enables our model to perform significantly better on unseen object instances or even on novel categories, as demonstrated in the experiments. 
Second, the distillation through rendering enforces \textit{consistent} 3D scene representation, leading to very similar features for the same object geometries. 
This enables the model to infer robust local \textit{correspondences} (see \cref{fig:ablation_correspondence}), which aids in fine-grained object matching.  %

\subsection{Object-centric Feature Extraction}
\label{sec:object_centric}
Having the enhanced 3D feature volume representation, we aim to localize each object instance in the scene and represent each with a 3D feature volume such that the instances can be compared to each other in the next step. 
To this end, we use the predicted density volume $\bm{V}_\sigma$, obtained in \cref{sec:enhance}, to extract a coarse point-cloud structure of the scene by applying a threshold to $\bm{V}_\sigma$. 
Then, we employ DBScan~\cite{ester1996density}, a density-based clustering method to segment the foreground objects in the point cloud. 
This produces bounding box regions $\{\bm{b}_n\}_{n=1}^{N}$ corresponding to the objects in the scene. 
Finally, we apply RoI pooling~\cite{deng2021voxel} on the voxel representation $\bm{F}_{v}$ to extract object-centric feature volumes $\{\bm{z}_n\}_{n=1}^{N}$, each with a size of $d \times 8 \times 8 \times 8$. 
Note that we do not require ground truth box coordinates during training. Next, we pass them to the cross-instance matching module for fine-grained comparison.

\subsection{Cross-instance Matching} 
\label{sec:cross_matching}
{Our task requires comparing object instances with each other, which further demands geometric alignment for each pair. Since the relative pose of each instance with respect to the camera is unknown, we avoid explicit alignment and instead focus on finding local matches between them. 
To this end, for a given pair of object instances ($m$ and $n$), we compute the most similar local features between their feature volumes ($\bm{z}_n$ and $\bm{z}_m$) respectively: }
\begin{equation}
    \mathcal{C}^{nm}_k[i] = \texttt{top}_k[\beta(\bm{z}_n[i])^T\beta(\bm{z}_m[.])],
    \label{eq:corr}
\end{equation} 
where we denote the operation as $\mathcal{C}^{nm}_k$, which returns top-$k$ most relevant feature locations in $\bm{z}_m$ for the $i$-th voxel location in $\bm{z}_n$. This is achieved by first projecting each voxel feature utilizing the projector function $\beta$ (described in Sec.~\ref{sec:enhance} %
and then computing their voxel-level pairwise similarity followed by a $\texttt{top}_k$ operation. We use $\mathcal{C}^{nm}_k$ to perform sparse attention-based comparisons between multiple object volumes, as described next.

The query, key, and value embeddings are calculated using linear projections as:
\begin{equation}
\begin{split}
    \bm{Q}_n[i] = \bm{W}^{Q}{\bm{z}}_n[i], \quad \bm{K}_n[i] = \bm{W}^K{\bm{z}}_n[i], \\
    \quad  \bm{V}_n[i] = \bm{W}^{V}{\bm{z}}_n[i],
    \end{split}
\end{equation}
where the weights $\bm{W}^{Q}$, $\bm{W}^{K}$ and $\bm{W}^{V}$ are shared across all objects. Then, the attention is calculated as:
\begin{equation}
    \bar{\bm{z}}_n[i] = \sum_{\substack{m=1 \\ m \neq n}}^{N} \sum_{j \in \mathcal{C}_k^{nm}[i]}\text{softmax}\left(\frac{\bm{Q}_n[i]\bm{K}_m[j]}{\sqrt{d}}\right)\bm{V}_m[j].
\end{equation}
Unlike the vanilla self-attention module in standard transformers~\cite{dosovitskiy2020image} that uses all tokens for the attention computation, which is inefficient for our task and may introduce noisy interactions with irrelevant features, potentially degrading performance. To overcome this, we compute the sparse voxel attention only among geometrically corresponding voxel locations using $\mathcal{C}_k^{nm}$.

Next, the updated feature volume $\bar{\bm{z}}_n$ is passed through 3D CNN blocks to downsample by a factor of $1/8$, which is finally reshaped into a vector and fed to a 2-layer MLP outputting the final prediction $\hat{y}_n$. The classification loss is calculated as:
\begin{equation}
   \mathcal{L}^\text{bce} = \sum_{n=1}^{N} \ell_\text{bce}(\hat{y}_n,y_n),
   \label{eq:optceonly}
\end{equation} where $\ell_\text{bce}$ is the binary cross-entropy loss function. The total training loss of our framework is $\mathcal{L} = \mathcal{L}^{\text{bce}}+\mathcal{L}^r$. %

%% file: sec/4-experiment.tex
\section{Experiments}
\label{sec:exp}

\subsection{Datasets}
\textit{ToysAD-8K} comprises real-world objects from \textit{multiple} categories, enabling  evaluation of our model's generalization to unseen object categories. 
\textit{PartsAD-15K} includes a diverse collection of mechanical parts with arbitrary shapes, eliminating class-level inductive biases present in common objects (\eg a sheep with three legs are easy to be recognized as anomaly without comparison).
Both datasets feature fine-grained anomaly instances inspired by real-world applications in inspection and quality control applications. 
Scenes are rendered with diverse backgrounds, illuminations, and camera viewpoints using photo-realistic ray tracing~\cite{blender}. 
Next, we discuss the data generation for both datasets.

\noindent \textbf{ToysAD-8K.} We start with a subset of $1050$ shapes from the Toys4K dataset~\cite{stojanov2021using}, covering a wide range of objects across $51$ categories.
Anomalies are automatically generated for each 3D shape by applying various deformations to both the geometry and texture.
These include realistic cracks and fractures using \cite{sellan2023breaking}, applying random geometric deformations~\cite{blender} like bumps, bends, and twists, as well as randomly translating, rotating, and swapping materials in different parts of the shapes.
This process results in a total of  $2345$ anomaly shapes. 

Each scene is generated by randomly selecting a set of objects, including both the normal and anomalous variants of the same instance. 
While most scenes contain more normal objects than anomalies, some scenes includes only normal instances. 
Object resting poses are determined through rigid body simulation~\cite{baraff2001physically} where objects are dropped onto the floor. %
Objects are scaled and placed randomly in the scene while avoiding collisions 
We generated a total of $8K$ scenes. Each scene consists of $3$-$6$ objects rendered in $20$ views.
For the training set, we randomly select $5K$ scenes from $39$ categories. We build two disjoint test sets. The first one (\textit{seen}) contains $1K$ scenes from the seen categories but with unseen object instances. The second one (\textit{unseen}) contains $2K$ scenes from the rest of the $12$ novel categories.

\noindent \textbf{PartsAD-15K.} We use a subset of the ABC dataset~\cite{Koch_2019_CVPR} that consists of $4200$ shapes. Anomalies are generated following the same strategy outlined above. 
Additionally, for each shape, we sample geometrically close instances from the dataset and assign them as anomalies to use in the same scene. This results in high-quality anomalies that closely resemble their normal counterparts in overall geometry but differ in subtle shape variations, making them anomalous in the context of the normal ones. 
In total, we generate $10,203$ anomaly shapes and create $15K$ scenes, each consisting of $3$ to $12$ objects rendered from $20$ different viewpoints. The dataset is split into a $12K$ training set and the remaining $3K$ scenes (unseen shapes) as the test set.  %

\noindent \textbf{Anomaly generation and checking.}
Our anomaly generation process is fully automated with several quality checks to ensure the anomalies are realistic. 
For example, if a part detaches from the main body during positional or rotational deformations, the anomaly is discarded and regenerated with adjusted parameters. Similarly, if removing a part makes the shape disconnected (\eg teddy bear with a floating hand after removing an arm)%
, we discard the anomaly and try a different part. 
For fracture anomalies, if a particular fracture removes more than $90\%$ or less than $10\%$ of an object, it is rejected and regenerated. 
We also ensure that the anomalous region of an object is visible from at least one viewpoint.

\begin{figure*}[!htb]
\begin{center}
   \includegraphics[width=0.99\linewidth]{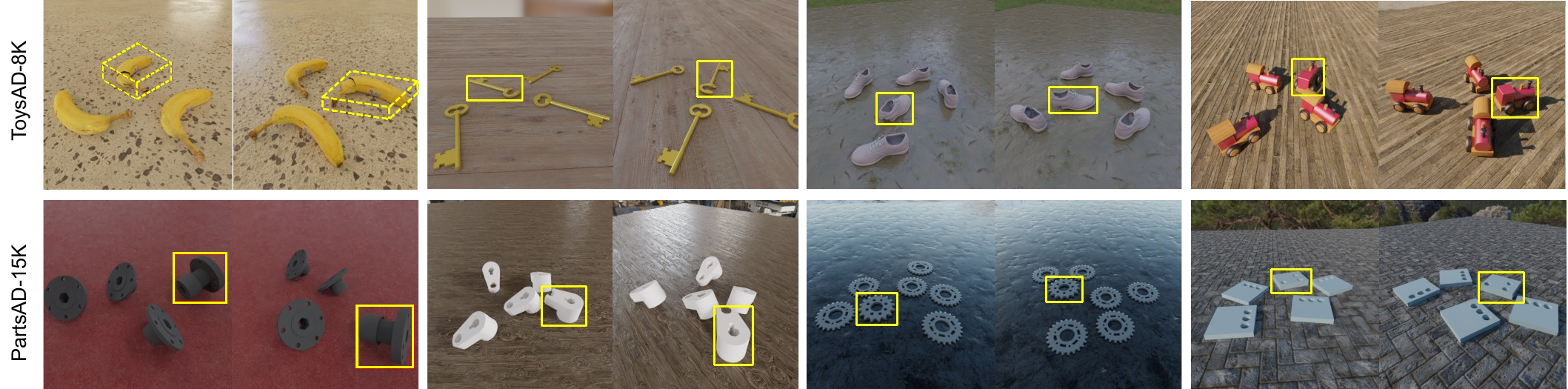}
\end{center}
\vspace{-0.5cm}
\caption{AD qualitative results on the \textit{unseen} test categories of ToysAD-8K (top row) and the test set of PartsAD-15K (bottom row) using our proposed framework. Due to limited space, two views are shown per scene. The model prediction is shown with a yellow bounding box: a 3D box for the first example (banana) and a projected box for the others for simplicity. Our model successfully predicts the correct object in all cases shown above.} %
\label{fig:results}
\end{figure*}

\begin{table*}[!htb]
\centering
\caption{
\textbf{Quantitative results.} We compare our method to three related works in two datasets and report the results in terms of anomaly detection AUC and accuracy. %
}
\vspace{-2mm}
\scalebox{0.90}{
\vspace{1em}
\begin{tabular}{lccccccccc}
    \toprule
    \multirow{2}{*}{Datasets} & \multicolumn{2}{c}{Recon-Recog} & \multicolumn{2}{c}{ImVoxelNet~\cite{rukhovich2022imvoxelnet}} & \multicolumn{2}{c}{DETR3D~\cite{wang2022detr3d}} & \multicolumn{2}{c}{\bf{Ours}} \\
    \cmidrule(lr){2-3} \cmidrule(lr){4-5} \cmidrule(lr){6-7} \cmidrule(lr){8-9}
	& AUC & Accuracy & AUC & Accuracy & AUC & Accuracy & AUC & Accuracy \\
    \cmidrule(lr){1-1} \cmidrule(lr){2-3} \cmidrule(lr){4-5} \cmidrule(lr){6-7}  \cmidrule(lr){8-9}
    ToysAD-8K-\textit{Seen}  & 73.45 & 60.48 & 78.13 & 65.55  & 79.16 & 67.37 & \bf{91.78} & \bf{83.21} \\
    ToysAD-8K-\textit{Unseen}   & 72.86 & 58.12  &   73.19  & 60.12 & 74.60 & 62.98 &   \bf{89.15} & \bf{81.57}\\
    \cmidrule(lr){1-1} \cmidrule(lr){2-3} \cmidrule(lr){4-5} \cmidrule(lr){6-7}  \cmidrule(lr){8-9}
    PartsAD-15K & 72.78 & 61.34  &  72.80 & 64.34 & 74.49 & 65.11 & \bf{86.12} & \bf{79.68} \\
    \bottomrule
\end{tabular}
}
\label{tab:quantitative}
\vspace{-2mm}
\end{table*}

\subsection{Implementation details}
\label{subsec:impl_details}
We use a ResNet50-FPN~\cite{lin2017feature} as the 2D encoder backbone and a four-scale encoder-decoder-based 3D CNN~\cite{murez2020atlas} as the 3D backbone.
For input, we use $M$=$5$ views, each with a resolution of $256\times 256$, though our model can accept a different $M$ during inference. 
During training, we use $2M$ views, divided into two sets of $M$ views. 
One set is used to build the neural feature volume, while the other set's camera views are used to render the results, and vice versa. 

The 3D volume $\bm{F}_v$ contains $96 \times 96 \times 16$ voxels with a voxel size of $4cm$. 
We sample $128$ points per ray for rendering, and the rendered features ($\hat{\Phi}_t$) have a spatial dimension of $32 \times 32$ (\ie $h_f \times w_f$). %
The ground truth segmentation mask of the input image ($I_{t\sigma}$ in \cref{eq:renderer}) is only used during training.
The threshold applied to the density volume $\bm{V}_\sigma$ is set to be $0.2$ and the DBScan algorithm is run with its default parameters.

We resize the teacher's (DINOv2) features to the same spatial dimension for loss computation.
These features are precomputed for all scenes using the publicly available weights, which are not updated during distillation. 
These DINOv2 features are then reduced to $d_f$=$128$ channel dimension using PCA before distillation.
We employ three sparse voxel attention blocks, and each applies 8-headed attention with $k$=$20$. 
The network is first pretrained for $50$ epochs with only the reconstruction loss to build a reliable initialized 3D feature volume, then trained end-to-end with both the reconstruction and the binary classification losses for another $50$ epochs. 
The batch size is $4$, and we use the Adam optimizer with a learning rate $2\times10^{-5}$.
The run-time of our method is $65$ms on a single A40 GPU for a typical scene with 5 views as input.

\subsection{Baseline Comparisons}

We evaluate the anomaly classification results using two metrics -- the area under the ROC curve (AUC) and accuracy.
A prediction is considered correct if the bounding box IoU is greater than 0.5 and the anomaly classification is correct.
We do not use a separate localization metric since our bounding box estimations are highly accurate. %
These metrics are calculated object-wise and then averaged across all test scenes.

Since no prior work exists for this task, we compare our method with two relevant competitive baselines: a reconstruction-based baseline and two multi-view 3D object detection methods in ~\cref{tab:quantitative}.
For the reconstruction-based baseline, we first employ COLMAP~\cite{schoenberger2016mvs} to generate a point cloud reconstruction of the multi-object scene environment. 
We use default dense reconstruction parameters but utilize the provided ground truth camera matrices. 
Given that 5 views are insufficient, we use a total of 20 views to reconstruct the scene. 
We extract point clouds from the reconstructed scenes for each object, and train a Siamese-style network to obtain their pairwise similarity. 
We use DGCNN~\cite{wang2019dynamic} (pretrained on ShapeNet~\cite{chang2015shapenet}) as the point cloud feature extractor and the triplet loss~\cite{hoffer2015deep} to supervise the network. 
A voting strategy is then applied to aggregate the pairwise distances of all objects in the scene and classify the objects. 
However, this method's performance is highly sensitive to the reconstruction quality, resulting in poor performance on both datasets. 

Next, we adapt two multi-view object detection frameworks, ImVoxelNet~\cite{rukhovich2022imvoxelnet} and DETR3D~\cite{wang2022detr3d}, to locate and classify each object in the scene as either anomalous or normal. ImVoxelNet constructs a voxel representation using a 2D-3D projection similar to ours, followed by a 3D detection head~\cite{lang2019pointpillars} for the final prediction. 
DETR3D, a transformer-based design, uses the set prediction loss~\cite{carion2020end} for end-to-end detection without non-maximum suppression. Both models are trained on our datasets and their performance is evaluated object-wise based on the output of the classification head, similar to ours. %
After examining the qualitative results (see supplementary), we observe that although both methods perform well for large cracks or fractures, they struggle when intra-group comparison is necessary. %
This is because they tend to memorize certain anomaly types and fail to capture the cross-object relations in the scene. 

Our method significantly outperforms all baselines on both datasets (see \cref{tab:quantitative}). 
This highlights the effectiveness of our architecture, which is designed to match corresponding regions across objects. 
Notably, our model exhibits a smaller performance drop on the \textit{unseen} set, thanks to the robust 3D representation that generalizes to novel categories.  Qualitative results of our method are shown in \cref{fig:results}

\subsection{Ablations and Model Analysis}

In this section, we present an ablation study of our model, analyze its robustness, and evaluate its performance on real scenes. 
Unless stated otherwise, all experiments are conducted on the ToysAD-8K \textit{unseen} set using 5 input views.

\noindent \textbf{Ablation of architecture design.} 
We evaluate the main components of our model through an ablation study and report the results in \cref{tab:ablation}. 
All variants include the 3D feature fusion module and are optimized at least for the image reconstruction loss, which is essential for constructing scene geometry. 
Variant \texttt{A} directly maps object-centric features to their binary labels using an MLP without comparing them.
In contrast, variant \texttt{B} applies standard attention layers to the object-centric features to learn cross-object correlations. 
Despite the attention layers, \texttt{B} only shows a slight improvement (+2.1\% AUC) over \texttt{A}.

\begin{table}[!htb]
\centering
\caption{Ablation Results on ToysAD-8K.}
\vspace{-2mm}
\label{tab:ablation}
\scalebox{0.85}{
\begin{tabular}{lcc}
	\toprule
	Variants & AUC & Accuracy\\
	\midrule
	\texttt{\textit{A}}: baseline &  79.13  & 66.78 \\ 
        \texttt{\textit{B}}: \texttt{\textit{A}} + vanilla attention & 81.24  & 68.20 \\ 
	\texttt{\textit{C}}: \texttt{\textit{B}} + feature distillation &  87.05 & 79.56 \\ 
        \midrule
        \texttt{Final}: \texttt{\textit{C}} + sparse voxel attention &  \bf{89.15} & \bf{81.57} \\ 

	\bottomrule
\end{tabular}
}
\end{table}

\begin{figure*}[t!]
\begin{center}
   \includegraphics[width=1\linewidth]{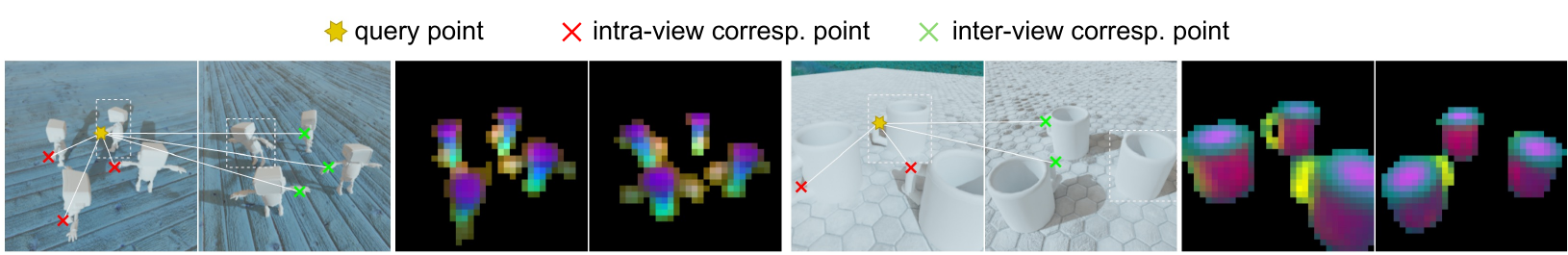}
\end{center}
\vspace{-0.7cm}
\caption{Correspondences are obtained in 3D using the \textit{neural feature field} $\bm{V}_f$ and projected onto 2D views via camera matrices for visualization. The feature field is rendered at respective viewpoints, with the first 3 PCA components mapped to different color channels. %
}
\label{fig:ablation_correspondence}
 \vspace{-0.4cm}
\end{figure*}

We then introduce DINOv2 feature distillation (variant C), which significantly improves performance by 5.8\% AUC and 11.3\% accuracy. 
This demonstrates the importance of the part-aware 3D representation and correspondences for our task (see \cref{fig:ablation_correspondence}).
Our final design achieves further performance gain by using sparse voxel attention, which focuses on the top-$k$ most relevant features. This leverages robust correspondences learned through DINOv2 feature distillation, effectively eliminating noisy correlations and directing attention solely to corresponding object regions.

\begin{figure}[h!]
\begin{center}
   \includegraphics[width=0.9\linewidth]{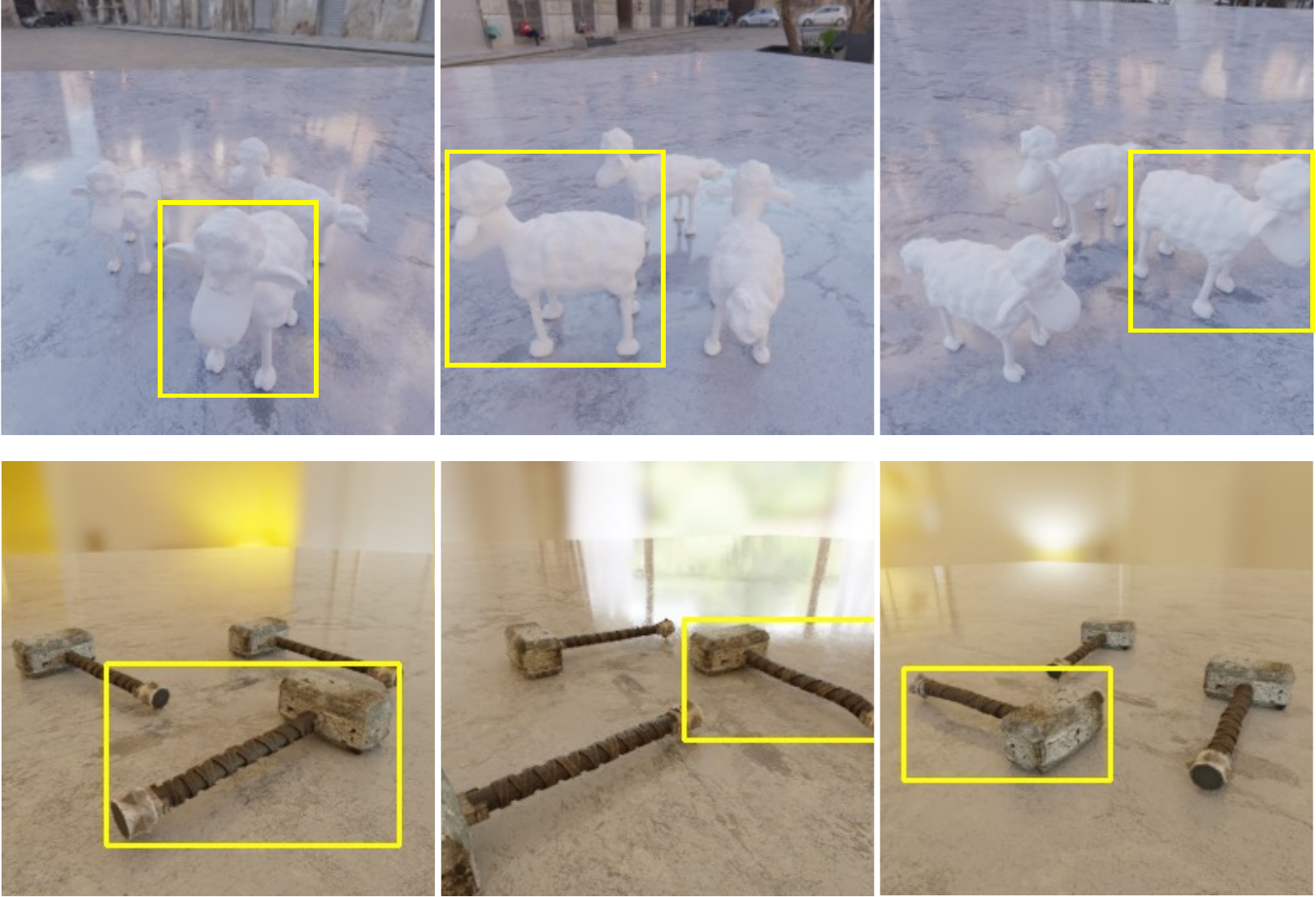}
\end{center}
\vspace{-0.4cm}
\caption{Resolving occlusion and 3D ambiguity  using multi-view images. The anomaly `sheep' in \textit{top} has a missing tail (only visible in the 2nd view due to occlusion), and the `hammer' handle in  \textit{bottom} is bent (only apparent from the 2nd view-angle due to 3D ambiguities).} %
\label{fig:occlusion}
 \vspace{-0.4cm}
\end{figure}

\begin{figure}[h!]
\begin{center}
   \includegraphics[width=1\linewidth]{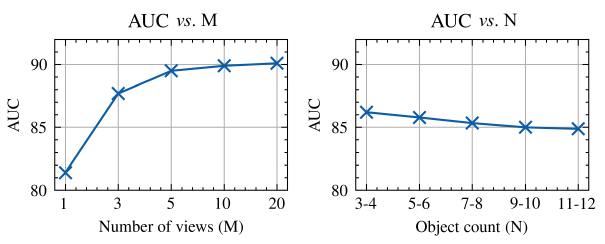}
\end{center}
\vspace{-0.6cm}
\caption{Impact of the number of views (left) and object count (right) on model performance. %
}
\label{fig:plot}
\vspace{-0.2cm}
\end{figure}

\noindent \textbf{Robustness.} %
Our method demonstrates robustness to occlusions by effectively utilizing input from multiple viewpoints. 
As shown in \cref{fig:occlusion}, our model accurately identifies the anomalous region even when parts of the object are occluded in some views. 
In \cref{fig:plot} (left), we show how well our model can perform with varying numbers of input views during testing. 
Trained with 5 views, the model is evaluated using 1, 3, 5, 10, and 20 views. 
The results clearly show that our model performs well with only 3-5 views, while additional views further enhance the performance.

We analyze the impact of object count on model performance (see \cref{fig:plot} (right)) using the PartsAD-15K dataset. 
As the number of objects in a scene increases, occlusion and lower resolution for individual objects typically occur. %
However, the performance drop between the two extremes is minimal ($<$ 1.5\% AUC) as shown in the figure. 
We also assess the model's ability to generalize to scenes with more objects than it encountered during training. 
To this end, we trained our model on scenes with 3-7 objects and tested on two sets: one with 3-7 objects (AUC: 86.75) and another with 8-12 objects (AUC: 85.10). These results demonstrate our model's adaptability to varying object counts. %

\begin{figure}[t!]
\begin{center}
   \includegraphics[width=1\linewidth]{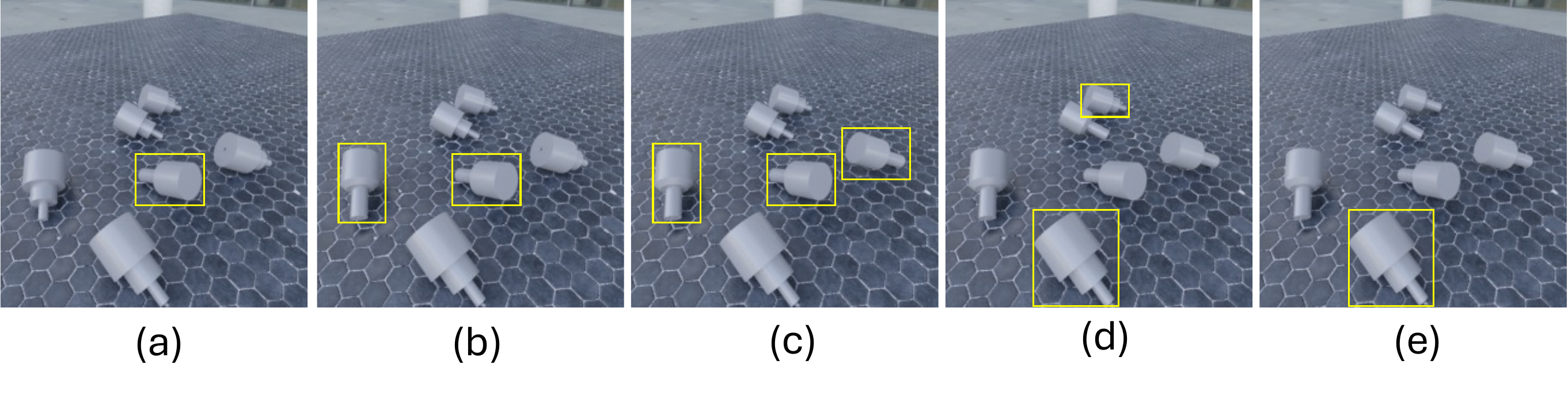}
\end{center}
\vspace{-0.8cm}
\caption{We create five scenes (a-e) with two similar objects, varying their counts per scene while maintaining a constant total. Our model correctly identifies minorities as anomalies in all cases except (c), where an equal number of objects creates ambiguity. Our model still selects one group. The yellow box indicates the predicted anomaly object.}
\label{fig:ratio}
 \vspace{-0.1cm}
\end{figure}

In another experiment, we assess our model's performances when varying the ratio of normal to anomalous instances during testing as shown in \cref{fig:ratio}. 
We create five scenes (a-e) by using two geometrically similar objects \texttt{ob1} and \texttt{ob2}, %
and gradually introduce more instances of \texttt{ob1} (1 to 5 respectively) while maintaining a total of six objects. 
For example, in scene (a), \texttt{ob1} appears only once, making it an anomaly. Similarly, in scene (e), \texttt{ob2} appears only once, hence considered an anomaly. We utilize a consistent background across all scenes to ensure uniformity. As shown in the figure, our model correctly classifies the anomalies in each scene. We note scene (c) presents an ambiguous case, where both objects appear in equal numbers. 
Despite this, our model is able to separate the two groups.

\begin{figure}[!t]
\centering
 \includegraphics[width=0.9\linewidth]{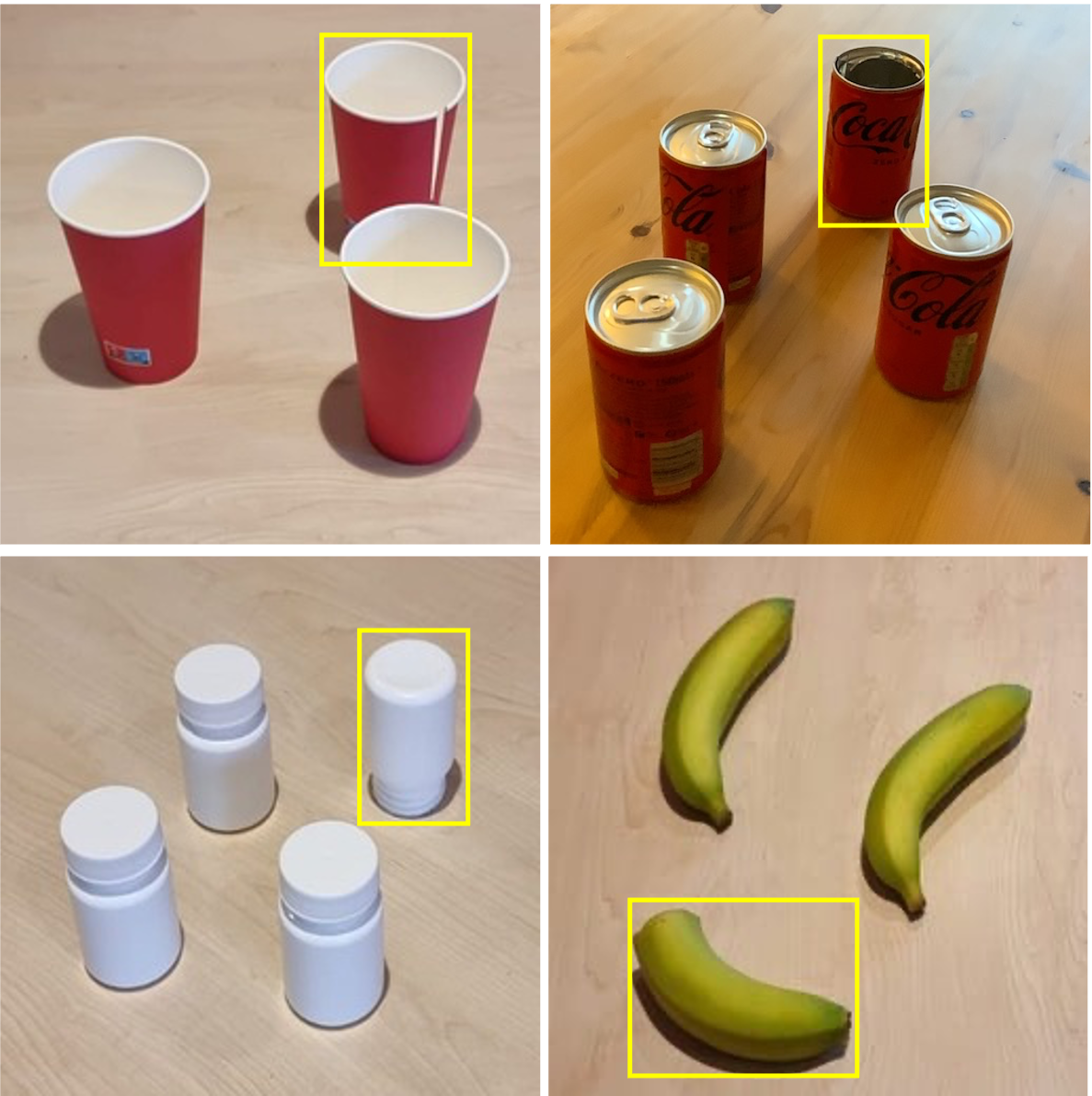}
\vspace{-2mm}
\caption{Four real-world scenes are tested using our method and it can successfully detect all anomalous instances. Five views are used for testing, but only one view per scene is shown.}
\label{fig:realworld}
 \vspace{-0.6cm}
\end{figure}

\noindent \textbf{Real world testing.} 
We apply our model, trained on the synthetic dataset, to a small set of real test scenes (a tabletop with three cups, four coca-cola cans, four medicine bottles, or three bananas). 
Each scene is captured in an indoor environment with adequate lighting using a 3D scanning software~\cite{Polycam}. We obtain a set of input views of the scenes with globally optimized camera intrinsic and extrinsic parameters, which are then fed into our trained model. The results are shown in \cref{fig:realworld}. Our model is able to correctly distinguish the anomaly objects in all four scenes.

\begin{figure}[t!]
\vspace{-0.1cm}
\begin{center}
   \includegraphics[width=1\linewidth]{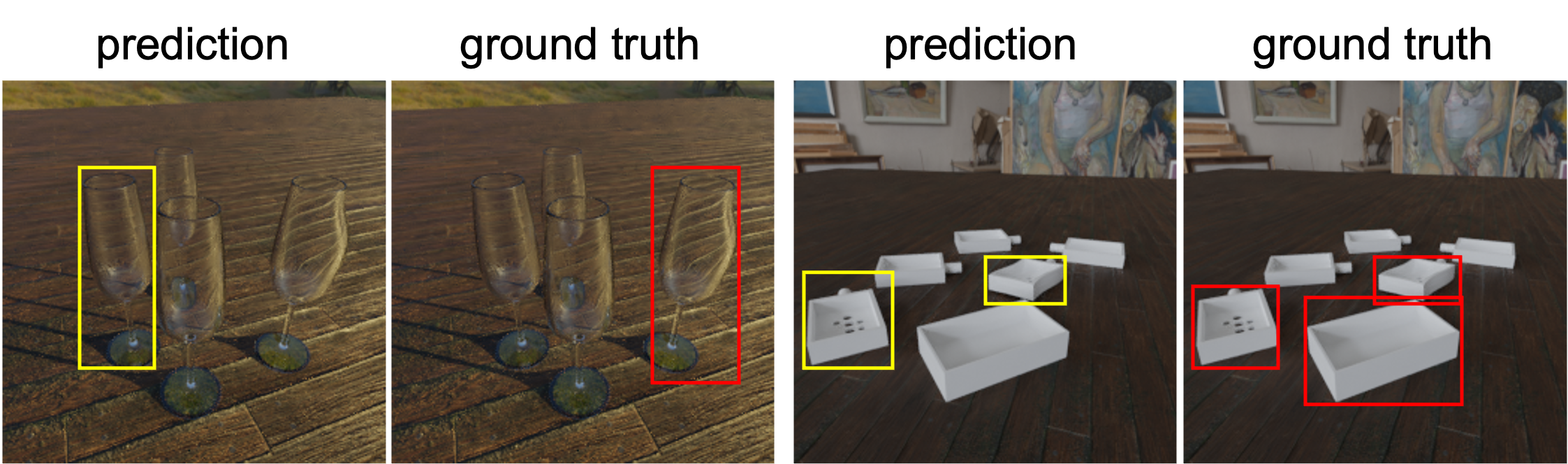}
\end{center}
\vspace{-0.5cm}
\caption{Failure cases. We use yellow box for our model's prediction and red box for the ground truth. In the first example, our model incorrectly predicts an anomaly, while in the second, it correctly identifies two objects but misses the third anomaly.}
\label{fig:fail}
\vspace{-0.3cm}
\end{figure}

\noindent \textbf{Limitations.} Our benchmark and model come with a few limitations. 
First, this work targets specific anomalies common in manufacturing, which may limit its applicability to other types of anomalies found in different settings.
Given the challenges and cost of obtaining real damaged objects, our dataset primarily uses shapes of synthetic objects. 
Additionally, our model assumes rigid objects and does not account for articulations or deformations. It also relies on objects being not touching nor not fully occluded in the scene. 
Furthermore, when a scene contains mostly anomalies, the lack of regular instances for comparison (except in cases like fractures or cracks) can further complicate the detection. 
Finally, real-world scene testing may introduce factors like noisy camera poses and the gap between synthetic and real environments, which could impact the performance. \cref{fig:fail} illustrates some failure cases of our model. Most failure cases stem from obscure anomalies due to factors like low resolution, poor lighting, unresolved occlusions in cluttered scenes, limited multi-view images, or challenging objects (\eg transparent textures). 

\section{Conclusion}
We have introduced a novel AD problem inspired by real-world applications along with two new benchmarks.
The proposed task goes beyond the traditional AD setting and involves a cross study of objects in a scene from multiple viewpoints to identify the `odd-looking' instances.
We show that our model is robust to varying number of views and objects, and outperforms the baselines that do not consider cross-object correlations and part-aware representations.

%% file: sec/6-suppl.tex
\clearpage
\appendix
\setcounter{page}{1}

\setcounter{table}{0}
\renewcommand{\thetable}{A\arabic{table}}
\setcounter{figure}{0}
\renewcommand{\thefigure}{A\arabic{figure}}

\maketitlesupplementary

This supplementary is structured as follows: we present additional details of the proposed datasets in ~\cref{sec:extra_data}, training details in ~\cref{sec:train_details}, and additional results in ~\cref{sec:extra_qual}.

\section{Data Generation Details}
\label{sec:extra_data}

\begin{figure*}[!ht]
\begin{center}
   \includegraphics[width=1\linewidth]{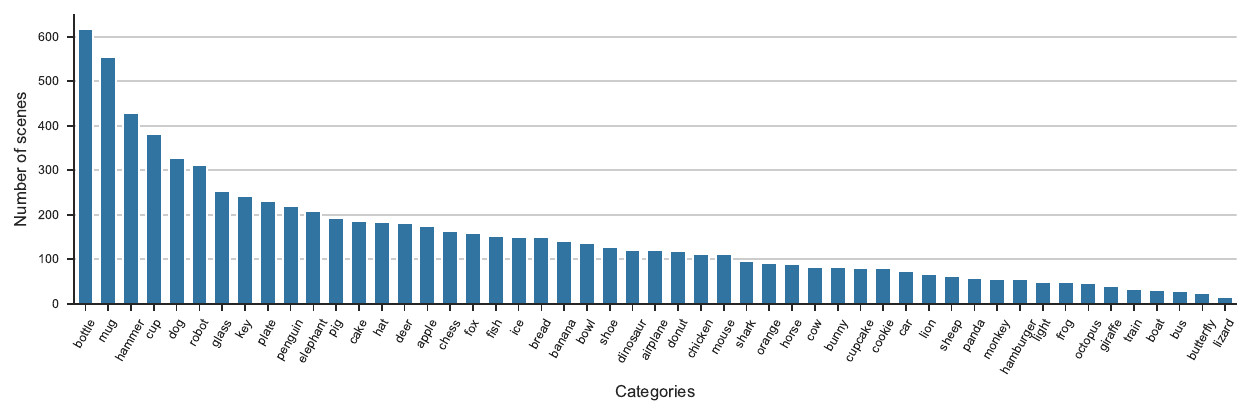}
\end{center}
\vspace{-7mm}
\caption{Distribution of object categories in ToysAD-8K dataset.}
 \label{fig:plot-dist}
\end{figure*}

Our proposed scene AD datasets, \emph{ToysAD-8K} and \emph{PartsAD-15K}, are built upon two publicly available 3D shape datasets: Toys4K~\cite{stojanov2021using} \textcolor{black}{(Creative Commons and royalty-free licenses)} and ABC~\cite{Koch_2019_CVPR} \textcolor{black}{(MIT license)}. For the ToysAD-8K, we selected 1,050 shapes from the Toys4K dataset, focusing on the most common real-world objects across 51 categories. A complete list of these categories is provided in~\cref{fig:plot-dist}, and the split of seen/unseen categories is shown in~\cref{tab:data_split}. On the other hand, PartsAD-15K is a non-categorical dataset. For this dataset, we randomly selected a subset of 4,200 shapes from the large-scale ABC. The test set of PartsAD-15K is comprised of unseen shapes. Given a 3D mesh model of an object, we automatically create anomalies by applying geometric deformations as described below.

\begin{table*}[!htb]
\centering
\vspace{-4mm}
\caption{Dataset composition of ToysAD-8K}
\label{tab:data_split}
\scalebox{0.85}{
\begin{tabular}{ll}
	\toprule
	 & Categories\\
	\midrule
	\multirow{3}{*}{Seen} &  dinosaur, fish, frog, monkey, light, lizard, orange, boat, dog, lion, pig, cookie, panda, chicken, \\  &  orange, ice, horse, car, airplane, cake, shark, donut, hat, cow, apple, bowl, hamburger, octopus, \\  &  giraffe, chess, bread, butterfly, cupcake,  bunny, elephant, fox, deer, bus, bottle  \\ 
 \midrule
        Unseen & mug, plate, robot, glass, sheep, shoe, train, banana, cup, key, penguin, hammer\\ 

	\bottomrule
\end{tabular}
}
\end{table*}

\begin{figure*}
\begin{center}
   \includegraphics[width=0.66\linewidth]{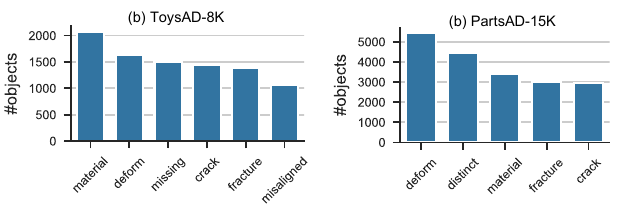}
\end{center}
\vspace{-0.6cm}
\caption{Distribution of anomaly types in the proposed ToysAD-8K and PartsAD-15 datasets.}
 \vspace{-0.cm}
 \label{fig:plot-types}
\end{figure*}

\begin{figure}
\begin{center}
   \includegraphics[width=1\linewidth]{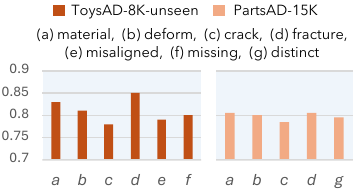}
\end{center}
\vspace{-0.6cm}
\caption{Anomaly detection performance (Accuracy) by anomaly type on the ToysAD-8K and PartsAD-15 datasets.}
 \vspace{-0.cm}
 \label{fig:accuracy-plot-types}
\end{figure}

\noindent \textbf{Different types of Anomalies.}  We consider the following anomaly types: cracks, fractures, geometric deformations (\eg bumps, bends, and twists), misaligned parts (\eg translation and rotation), material mismatch, and missing parts. 
\cref{fig:plot-types} depicts the distribution of different anomaly types in our proposed dataset. 
Before applying deformations, we normalize the mesh vertices to $[-1,1]$. 
Then we synthetically generate realistic cracks and fractures following \cite{sellan2023breaking} to mimic the shape’s most geometrically natural breaking patterns. For geometric deformations, we use Blender's~\cite{blender} Simple Deform and Hook modifier to create global (\eg bending and twisting) and local deformations (\eg surface bumps), respectively. 
For translation, we randomly translate the vertices of a random part by an offset. The translation offset for each axis is randomly sampled from a uniform distribution of range $[0.04, 0.08]$. 
For rotational anomalies, we apply a 3D rotational transformation to a random part. The rotation matrix is formed using a random rotation axis and a radian angle sampled from a uniform distribution of range $[0.2, 0.4]$. The center of rotation is set to a fixed point at one of the connecting points between the anomalous part and the main body of the object. We also randomly change the texture of a part or region to create a material mismatch anomaly or completely remove a part to simulate a missing parts anomaly. Additionally, for PartsAD-15K, we use a different but geometrically similar instance in the same scene as an anomaly (referred to as `distinct' in \cref{fig:plot-types}). 
To retrieve a geometrically similar shape, we use feature-based KNN clustering. Specifically, we extract DINOv2 features from multi-view images (rendered from multiple fixed viewpoints) of the 3D shapes, concatenate these features, and use the resulting concatenated features to build the KNN cluster. Then, we retrieve a similar 3D shape by querying the KNN cluster. To ensure the retrieved shape is geometrically similar, we calculate the Chamfer distance between the shapes and only accept a shape if the distance is below a certain threshold. 

\noindent \textbf{Scene construction.} For generating scenes, we use resulting anomalies along with their normal instances. 
A scene can have more than one anomaly of the same instance, or it may not have any anomaly at all. For randomized object placement, we first determine each object’s resting pose by simulating a rigid body drop using Blender~\cite{blender}. This simulation process is repeated for each object instance in the scene, including both normal and anomaly instances. Then, we place each object instance randomly using its obtained resting pose, while also ensuring that they do not collide with each other. We prevent collisions by maintaining a minimum distance between each pair of objects.

\noindent \textbf{Quality checks.} Our anomaly generation process is fully automated, with multiple quality checks to ensure reliable samples. 
For positional or rotational anomaly creation, we discard and regenerate any instance where a part detaches from the main body during deformation. Similarly, if removing a part results in an impractical shape, we try a different part instead. 
For fracture anomalies, we discard the sample where the fracture removes more than 90\% or less than 10\% of an object, regenerating a new fracture in these cases. Finally, we ensure the anomalous region of an object is visible from at least one viewpoint.

\noindent \textbf{Assets.}  To generate a realistic scene environment, we use PBR materials~\cite{thai2024low} for floors and HDRI environment maps~\cite{zaal2021polyhaven} for image-based lighting to illuminate scenes. We randomly select a pair of PBR material and HDRI environment maps from the assets to randomize the scene background. 
We employed Blender 2.93~\cite{blender} with Cycles ray-tracing renderer for photo-realistic rendering. \textcolor{black}{Blender 2.93 is released under the GNU General Public License (GPL, or “free software”), and the PBR and HDRI maps are released under the CC0 license.}

\noindent \textbf{Rendering and view selection.} 
We render each scene from various viewpoints sampled from a hemisphere around the scene origin. The viewpoint is parameterized in spherical coordinates where azimuth values are sampled uniformly over [0, $2\pi$) and elevation values are uniformly sampled in [$\pi/9$, $2\pi/9$]. We adjust the radius in the range [1.5, 2.5] to ensure that all the objects in the scene are visible (even if partially) in the rendered view.

Our framework can easily be trained with real-world manufacturing scene environments. Our model relies solely on 2D supervision, making the data collection process much easier. We also do not need precise annotation of 3D bounding boxes as they are not used during training. For annotating instance-wise anomaly labels, we can use 2D bounding boxes, which can be projected in 3D using Visual Hull \cite{kutulakos2000theory}, then used for coarse localization.

\section{Training Details}
\label{sec:train_details}
We train our model in two stages. In the first stage, we train it with just image and feature reconstruction losses. In the second stage, we train the model end-to-end with both reconstruction and binary classification losses. All the experiments are performed on a single NVIDIA A40 GPU with a batch size of 4, utilizing 28GB of GPU memory. The first stage takes 36 hours to complete, followed by an additional 24 hours for the second stage.

\textcolor{black}{The ablation experiments are conducted on the same workstation with the same GPU by removing one or a few core components from the full method. Specifically, variant methods \emph{A} and \emph{B} take $36$ hours to train the first stage, while only taking $14$ hours to train the second stage. Regarding variant method \emph{C}, it takes similar $36$ and $24$ hours for the two stages as in the full method.}

\textcolor{black}{We compare our method with Recons-Recog~\cite{schoenberger2016mvs}, ImVoxelNet~\cite{rukhovich2022imvoxelnet}, and DETR3D~\cite{wang2022detr3d}. For the Recons-Recog approach, we use DGCNN~\cite{wang2019dynamic} as a feature extractor.}
\textcolor{black}{In terms of training time of these baseline methods, Recons-Recog takes $10$ hours to train, ImVoxelNet takes $24$ hours to train, and DETR3D takes $2$ days to converge.}

\begin{figure*}
\begin{center}
   \includegraphics[width=1\linewidth]{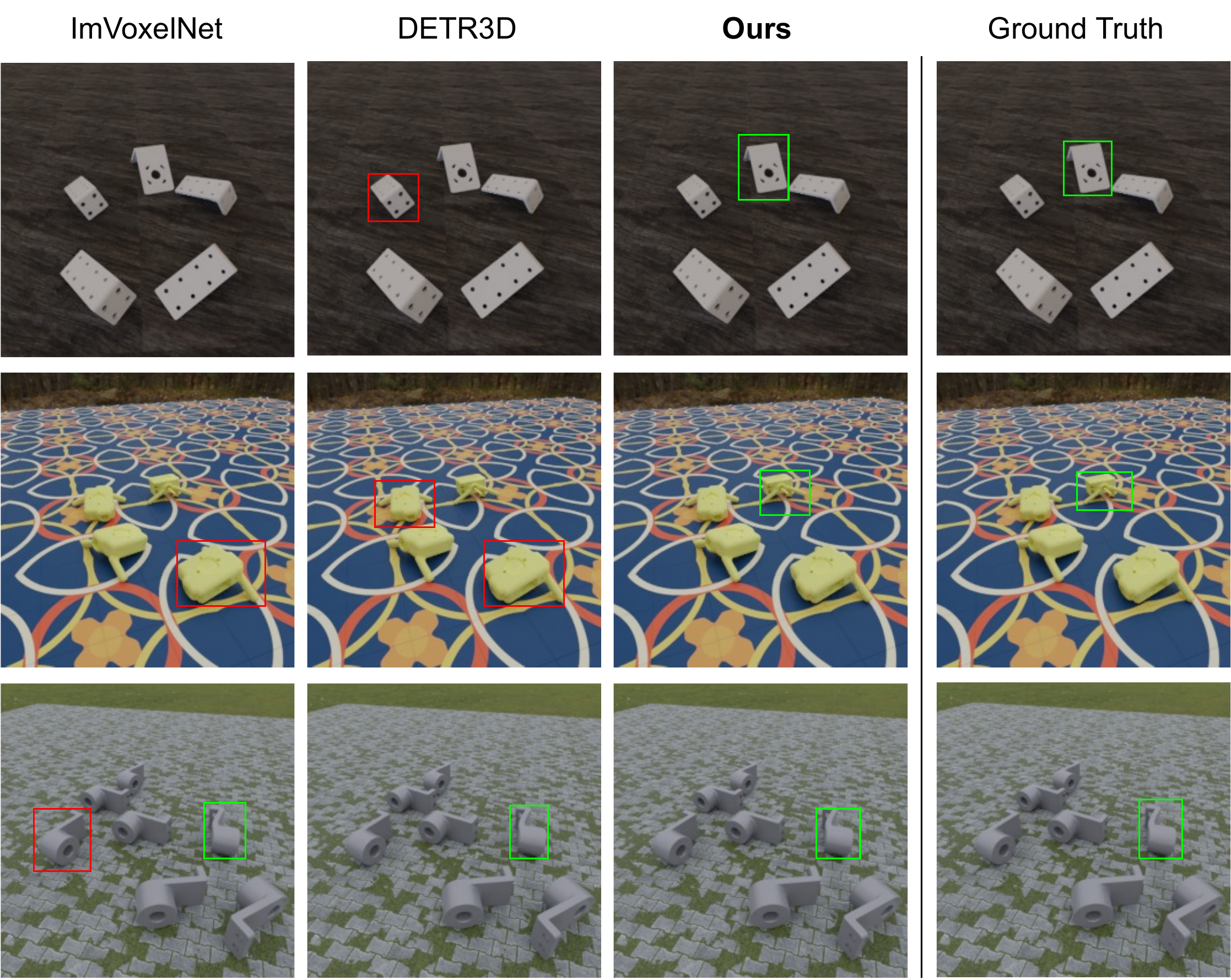}
\end{center}
\vspace{-0.25cm}
\caption{We qualitatively compare our method with two baselines on the PartsAD-15K dataset.  ImVoxelNet failed to detect the anomalous object in the first example. We use the red box for wrong predictions and the green box for correct predictions.}
 \vspace{-0.2cm}
 \label{fig:compare-baseline}
\end{figure*}

\section{Additional Results}
\label{sec:extra_qual}
\begin{figure*}[!htb]
\begin{center}
   \includegraphics[width=1\linewidth]{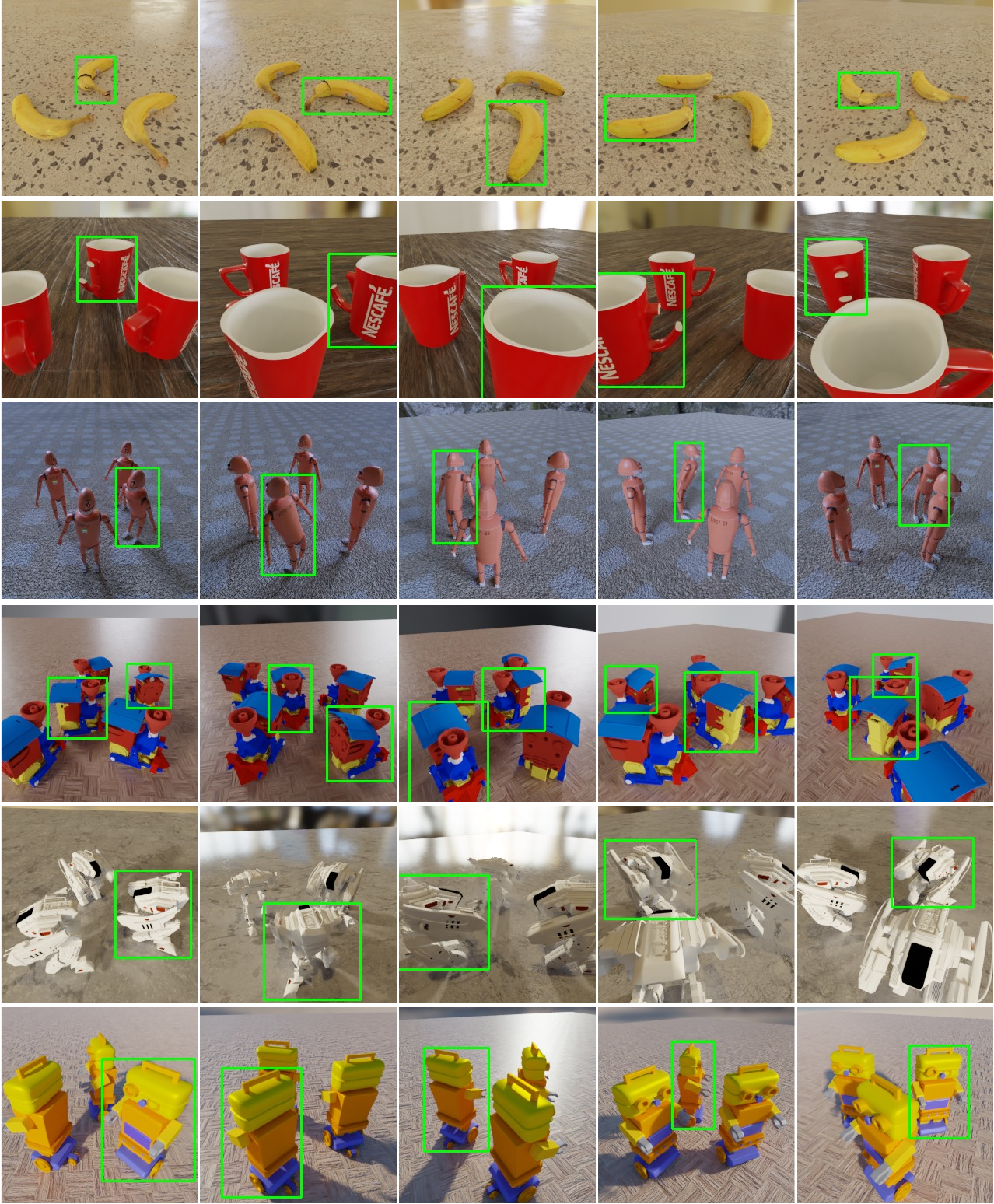}
\end{center}
\vspace{-0.25cm}
\caption{Additional results on the \textit{unseen} set of ToysAD-8K dataset. Each row shows multiple views of the input scene and the green box denotes our model's prediction. For rows 1 to 3, the anomalies are easy to spot and self-explanatory. In row 4, there are two anomalies: one with broken outer parts at the back (see view 5), and another with a tilted roof (see views 1 and 2). For row 5, one leg is broken (see view 2). In row 6, the eye is missing (see view 5).}
 \vspace{-0.2cm}
 \label{fig:more_toys}
\end{figure*}

\begin{figure*}[!htb]
\begin{center}
   \includegraphics[width=1\linewidth]{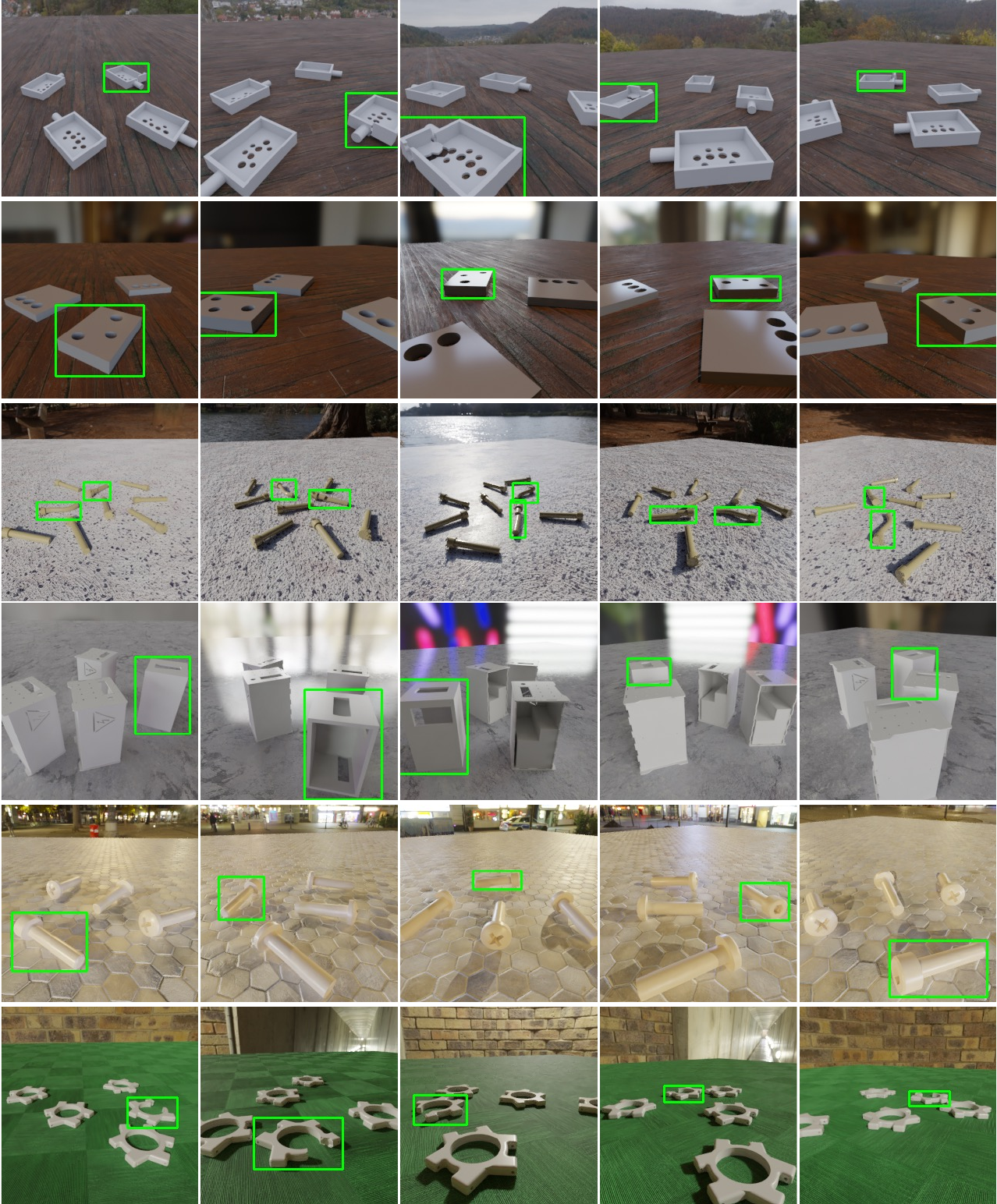}
\end{center}
\vspace{-0.25cm}
\caption{Additional results on the PartsAD-15K dataset. Each row shows multiple views of the input scene and the green box denotes our model's prediction. }
 \vspace{-0.2cm}
 \label{fig:more_parts}
\end{figure*}

\cref{fig:compare-baseline} shows a qualitative comparison with two baseline methods: ImVoxelNet and DETR3D. Our method accurately predicts the anomalies, while the baselines perform poorly. In the first three examples, the baselines fail due to their inability to compare with other objects in the scene, which is necessary for correct predictions. In the last example, both DETR3D and our method predict correctly, while ImVoxelNet fails.
In \cref{fig:more_toys} and \cref{fig:more_parts}, we present additional qualitative results on ToysAD-8K and PartsAD-15K, respectively. We show a breakdown of accuracy across different anomaly types in~\cref{fig:accuracy-plot-types} to interpret the model’s performance in various such scenarios.

%% file: main.bbl
\begin{thebibliography}{56}
\providecommand{\natexlab}[1]{#1}
\providecommand{\url}[1]{\texttt{#1}}
\expandafter\ifx\csname urlstyle\endcsname\relax
  \providecommand{\doi}[1]{doi: #1}\else
  \providecommand{\doi}{doi: \begingroup \urlstyle{rm}\Url}\fi

\bibitem[Pol(2022)]{Polycam}
Polycam.
\newblock https://github.com/PolyCam/polyform, 2022.

\bibitem[Ahmed and Courville(2020)]{ahmed2020detecting}
Faruk Ahmed and Aaron Courville.
\newblock Detecting semantic anomalies.
\newblock In \emph{AAAI}, 2020.

\bibitem[Banani et~al.(2024)Banani, Raj, Maninis, Kar, Li, Rubinstein, Sun, Guibas, Johnson, and Jampani]{banani2024probing}
Mohamed~El Banani, Amit Raj, Kevis-Kokitsi Maninis, Abhishek Kar, Yuanzhen Li, Michael Rubinstein, Deqing Sun, Leonidas Guibas, Justin Johnson, and Varun Jampani.
\newblock Probing the 3d awareness of visual foundation models.
\newblock In \emph{CVPR}, 2024.

\bibitem[Baraff(2001)]{baraff2001physically}
David Baraff.
\newblock Physically based modeling: Rigid body simulation.
\newblock In \emph{ACM SIGGRAPH}, 2001.

\bibitem[Bergmann et~al.(2019)Bergmann, Fauser, Sattlegger, and Steger]{bergmann2019mvtec}
Paul Bergmann, Michael Fauser, David Sattlegger, and Carsten Steger.
\newblock Mvtec ad--a comprehensive real-world dataset for unsupervised anomaly detection.
\newblock In \emph{CVPR}, 2019.

\bibitem[Bergmann et~al.(2021)Bergmann, Jin, Sattlegger, and Steger]{bergmann2021mvtec}
Paul Bergmann, Xin Jin, David Sattlegger, and Carsten Steger.
\newblock The mvtec 3d-ad dataset for unsupervised 3d anomaly detection and localization.
\newblock \emph{arXiv preprint arXiv:2112.09045}, 2021.

\bibitem[Bhunia et~al.(2024)Bhunia, Li, and Bilen]{bhunia2024looking}
Ankan Bhunia, Changjian Li, and Hakan Bilen.
\newblock Looking 3d: Anomaly detection with 2d-3d alignment.
\newblock In \emph{CVPR}, 2024.

\bibitem[Carion et~al.(2020)Carion, Massa, Synnaeve, Usunier, Kirillov, and Zagoruyko]{carion2020end}
Nicolas Carion, Francisco Massa, Gabriel Synnaeve, Nicolas Usunier, Alexander Kirillov, and Sergey Zagoruyko.
\newblock End-to-end object detection with transformers.
\newblock In \emph{ECCV}, 2020.

\bibitem[Carrera et~al.(2016)Carrera, Manganini, Boracchi, and Lanzarone]{carrera2016defect}
Diego Carrera, Fabio Manganini, Giacomo Boracchi, and Ettore Lanzarone.
\newblock Defect detection in sem images of nanofibrous materials.
\newblock In \emph{IEEE Transactions on Industrial Informatics}, 2016.

\bibitem[Chalapathy et~al.(2018)Chalapathy, Menon, and Chawla]{chalapathy2018anomaly}
Raghavendra Chalapathy, Aditya~Krishna Menon, and Sanjay Chawla.
\newblock Anomaly detection using one-class neural networks.
\newblock \emph{arXiv preprint arXiv:1802.06360}, 2018.

\bibitem[Chandola et~al.(2009)Chandola, Banerjee, and Kumar]{chandola2009anomaly}
Varun Chandola, Arindam Banerjee, and Vipin Kumar.
\newblock Anomaly detection: A survey.
\newblock In \emph{ACM Computing Surveys}, 2009.

\bibitem[Chang et~al.(2015)Chang, Funkhouser, Guibas, Hanrahan, Huang, Li, Savarese, Savva, Song, Su, et~al.]{chang2015shapenet}
Angel~X Chang, Thomas Funkhouser, Leonidas Guibas, Pat Hanrahan, Qixing Huang, Zimo Li, Silvio Savarese, Manolis Savva, Shuran Song, Hao Su, et~al.
\newblock Shapenet: An information-rich 3d model repository.
\newblock \emph{arXiv preprint arXiv:1512.03012}, 2015.

\bibitem[Chen et~al.(2021)Chen, Xu, Zhao, Zhang, Xiang, Yu, and Su]{chen2021mvsnerf}
Anpei Chen, Zexiang Xu, Fuqiang Zhao, Xiaoshuai Zhang, Fanbo Xiang, Jingyi Yu, and Hao Su.
\newblock Mvsnerf: Fast generalizable radiance field reconstruction from multi-view stereo.
\newblock In \emph{ICCV}, 2021.

\bibitem[Deecke et~al.(2021)Deecke, Ruff, Vandermeulen, and Bilen]{deecke2021transfer}
Lucas Deecke, Lukas Ruff, Robert~A Vandermeulen, and Hakan Bilen.
\newblock Transfer-based semantic anomaly detection.
\newblock In \emph{ICML}, 2021.

\bibitem[Deng et~al.(2021)Deng, Shi, Li, Zhou, Zhang, and Li]{deng2021voxel}
Jiajun Deng, Shaoshuai Shi, Peiwei Li, Wengang Zhou, Yanyong Zhang, and Houqiang Li.
\newblock Voxel r-cnn: Towards high performance voxel-based 3d object detection.
\newblock In \emph{AAAI}, 2021.

\bibitem[Ding et~al.(2022)Ding, Pang, and Shen]{ding2022catching}
Choubo Ding, Guansong Pang, and Chunhua Shen.
\newblock Catching both gray and black swans: Open-set supervised anomaly detection.
\newblock In \emph{CVPR}, 2022.

\bibitem[Dosovitskiy et~al.(2021)Dosovitskiy, Beyer, Kolesnikov, Weissenborn, Zhai, Unterthiner, Dehghani, Minderer, Heigold, Gelly, et~al.]{dosovitskiy2020image}
Alexey Dosovitskiy, Lucas Beyer, Alexander Kolesnikov, Dirk Weissenborn, Xiaohua Zhai, Thomas Unterthiner, Mostafa Dehghani, Matthias Minderer, Georg Heigold, Sylvain Gelly, et~al.
\newblock An image is worth 16x16 words: Transformers for image recognition at scale.
\newblock In \emph{ICLR}, 2021.

\bibitem[Ester et~al.(1996)Ester, Kriegel, Sander, and Xu]{ester1996density}
Martin Ester, Hans-Peter Kriegel, J{\"o}rg Sander, and Xiaowei Xu.
\newblock Density-based spatial clustering of applications with noise.
\newblock In \emph{ACM SIGKDD international conference on Knowledge discovery and data mining}, 1996.

\bibitem[Hoffer and Ailon(2015)]{hoffer2015deep}
Elad Hoffer and Nir Ailon.
\newblock Deep metric learning using triplet network.
\newblock In \emph{Similarity-Based Pattern Recognition: Third International Workshop, SIMBAD}, 2015.

\bibitem[Huang et~al.(2022)Huang, Guan, Jiang, Zhang, Spratling, and Wang]{huang2022registration}
Chaoqin Huang, Haoyan Guan, Aofan Jiang, Ya Zhang, Michael Spratling, and Yan-Feng Wang.
\newblock Registration based few-shot anomaly detection.
\newblock In \emph{ECCV}, 2022.

\bibitem[Jiang et~al.(2022)Jiang, Jiang, Grauman, and Zhu]{jiang2022few}
Hanwen Jiang, Zhenyu Jiang, Kristen Grauman, and Yuke Zhu.
\newblock Few-view object reconstruction with unknown categories and camera poses.
\newblock \emph{arXiv preprint arXiv:2212.04492}, 2022.

\bibitem[Kirillov et~al.(2023)Kirillov, Mintun, Ravi, Mao, Rolland, Gustafson, Xiao, Whitehead, Berg, Lo, et~al.]{kirillov2023segment}
Alexander Kirillov, Eric Mintun, Nikhila Ravi, Hanzi Mao, Chloe Rolland, Laura Gustafson, Tete Xiao, Spencer Whitehead, Alexander~C Berg, Wan-Yen Lo, et~al.
\newblock Segment anything.
\newblock In \emph{ICCV}, 2023.

\bibitem[Kobayashi et~al.(2022)Kobayashi, Matsumoto, and Sitzmann]{kobayashi2022decomposing}
Sosuke Kobayashi, Eiichi Matsumoto, and Vincent Sitzmann.
\newblock Decomposing nerf for editing via feature field distillation.
\newblock 2022.

\bibitem[Koch et~al.(2019)Koch, Matveev, Jiang, Williams, Artemov, Burnaev, Alexa, Zorin, and Panozzo]{Koch_2019_CVPR}
Sebastian Koch, Albert Matveev, Zhongshi Jiang, Francis Williams, Alexey Artemov, Evgeny Burnaev, Marc Alexa, Denis Zorin, and Daniele Panozzo.
\newblock Abc: A big cad model dataset for geometric deep learning.
\newblock In \emph{CVPR}, 2019.

\bibitem[Kutulakos and Seitz(2000)]{kutulakos2000theory}
Kiriakos~N Kutulakos and Steven~M Seitz.
\newblock A theory of shape by space carving.
\newblock In \emph{IJCV}, 2000.

\bibitem[Lang et~al.(2019)Lang, Vora, Caesar, Zhou, Yang, and Beijbom]{lang2019pointpillars}
Alex~H Lang, Sourabh Vora, Holger Caesar, Lubing Zhou, Jiong Yang, and Oscar Beijbom.
\newblock Pointpillars: Fast encoders for object detection from point clouds.
\newblock In \emph{CVPR}, 2019.

\bibitem[Lin et~al.(2017)Lin, Doll{\'a}r, Girshick, He, Hariharan, and Belongie]{lin2017feature}
Tsung-Yi Lin, Piotr Doll{\'a}r, Ross Girshick, Kaiming He, Bharath Hariharan, and Serge Belongie.
\newblock Feature pyramid networks for object detection.
\newblock In \emph{CVPR}, 2017.

\bibitem[Mildenhall et~al.(2021)Mildenhall, Srinivasan, Tancik, Barron, Ramamoorthi, and Ng]{mildenhall2021nerf}
Ben Mildenhall, Pratul~P Srinivasan, Matthew Tancik, Jonathan~T Barron, Ravi Ramamoorthi, and Ren Ng.
\newblock Nerf: Representing scenes as neural radiance fields for view synthesis.
\newblock In \emph{Communications of the ACM}, 2021.

\bibitem[Murez et~al.(2020)Murez, Van~As, Bartolozzi, Sinha, Badrinarayanan, and Rabinovich]{murez2020atlas}
Zak Murez, Tarrence Van~As, James Bartolozzi, Ayan Sinha, Vijay Badrinarayanan, and Andrew Rabinovich.
\newblock Atlas: End-to-end 3d scene reconstruction from posed images.
\newblock In \emph{ECCV}, 2020.

\bibitem[Oquab et~al.(2023)Oquab, Darcet, Moutakanni, Vo, Szafraniec, Khalidov, Fernandez, Haziza, Massa, El-Nouby, et~al.]{oquab2023dinov2}
Maxime Oquab, Timoth{\'e}e Darcet, Th{\'e}o Moutakanni, Huy Vo, Marc Szafraniec, Vasil Khalidov, Pierre Fernandez, Daniel Haziza, Francisco Massa, Alaaeldin El-Nouby, et~al.
\newblock Dinov2: Learning robust visual features without supervision.
\newblock \emph{arXiv preprint arXiv:2304.07193}, 2023.

\bibitem[Pang et~al.(2021)Pang, Shen, Cao, and Hengel]{pang2021deep}
Guansong Pang, Chunhua Shen, Longbing Cao, and Anton Van~Den Hengel.
\newblock Deep learning for anomaly detection: A review.
\newblock In \emph{ACM Computing Surveys}, 2021.

\bibitem[Radford et~al.(2021)Radford, Kim, Hallacy, Ramesh, Goh, Agarwal, Sastry, Askell, Mishkin, Clark, et~al.]{radford2021learning}
Alec Radford, Jong~Wook Kim, Chris Hallacy, Aditya Ramesh, Gabriel Goh, Sandhini Agarwal, Girish Sastry, Amanda Askell, Pamela Mishkin, Jack Clark, et~al.
\newblock Learning transferable visual models from natural language supervision.
\newblock In \emph{ICML}, 2021.

\bibitem[Ruff et~al.(2018)Ruff, Vandermeulen, Goernitz, Deecke, Siddiqui, Binder, M{\"u}ller, and Kloft]{ruff2018deep}
Lukas Ruff, Robert Vandermeulen, Nico Goernitz, Lucas Deecke, Shoaib~Ahmed Siddiqui, Alexander Binder, Emmanuel M{\"u}ller, and Marius Kloft.
\newblock Deep one-class classification.
\newblock In \emph{ICML}, 2018.

\bibitem[Rukhovich et~al.(2022)Rukhovich, Vorontsova, and Konushin]{rukhovich2022imvoxelnet}
Danila Rukhovich, Anna Vorontsova, and Anton Konushin.
\newblock Imvoxelnet: Image to voxels projection for monocular and multi-view general-purpose 3d object detection.
\newblock In \emph{WACV}, 2022.

\bibitem[Sch\"{o}nberger et~al.(2016)Sch\"{o}nberger, Zheng, Pollefeys, and Frahm]{schoenberger2016mvs}
Johannes~Lutz Sch\"{o}nberger, Enliang Zheng, Marc Pollefeys, and Jan-Michael Frahm.
\newblock Pixelwise view selection for unstructured multi-view stereo.
\newblock In \emph{ECCV}, 2016.

\bibitem[Sell{\'a}n et~al.(2023)Sell{\'a}n, Luong, Mattos Da~Silva, Ramakrishnan, Yang, and Jacobson]{sellan2023breaking}
Silvia Sell{\'a}n, Jack Luong, Leticia Mattos Da~Silva, Aravind Ramakrishnan, Yuchuan Yang, and Alec Jacobson.
\newblock Breaking good: Fracture modes for realtime destruction.
\newblock In \emph{ACM Transactions on Graphics}, 2023.

\bibitem[Shi et~al.(2021)Shi, Ye, Chen, Chen, Chen, and Kim]{shi2021geometry}
Xuepeng Shi, Qi Ye, Xiaozhi Chen, Chuangrong Chen, Zhixiang Chen, and Tae-Kyun Kim.
\newblock Geometry-based distance decomposition for monocular 3d object detection.
\newblock In \emph{ICCV}, 2021.

\bibitem[Stojanov et~al.(2021)Stojanov, Thai, and Rehg]{stojanov2021using}
Stefan Stojanov, Anh Thai, and James~M Rehg.
\newblock Using shape to categorize: Low-shot learning with an explicit shape bias.
\newblock In \emph{CVPR}, 2021.

\bibitem[Sun et~al.(2021)Sun, Xie, Chen, Zhou, and Bao]{sun2021neuralrecon}
Jiaming Sun, Yiming Xie, Linghao Chen, Xiaowei Zhou, and Hujun Bao.
\newblock Neuralrecon: Real-time coherent 3d reconstruction from monocular video.
\newblock In \emph{CVPR}, 2021.

\bibitem[Team(2022)]{blender}
Blender~Development Team.
\newblock Blender (version 3.1.0) [computer software].
\newblock \url{https://blender.org/}, 2022.

\bibitem[Thai et~al.(2024)Thai, Humayun, Stojanov, Huang, Boote, and Rehg]{thai2024low}
Anh Thai, Ahmad Humayun, Stefan Stojanov, Zixuan Huang, Bikram Boote, and James~M Rehg.
\newblock Low-shot object learning with mutual exclusivity bias.
\newblock In \emph{NeurIPS}, 2024.

\bibitem[Trevithick and Yang(2021)]{trevithick2020grf}
Alex Trevithick and Bo Yang.
\newblock Grf: Learning a general radiance field for 3d scene representation and rendering.
\newblock In \emph{ICCV}, 2021.

\bibitem[Tschernezki et~al.(2022)Tschernezki, Laina, Larlus, and Vedaldi]{tschernezki2022neural}
Vadim Tschernezki, Iro Laina, Diane Larlus, and Andrea Vedaldi.
\newblock Neural feature fusion fields: 3d distillation of self-supervised 2d image representations.
\newblock In \emph{3DV}, 2022.

\bibitem[Wang et~al.(2021)Wang, Wang, Genova, Srinivasan, Zhou, Barron, Martin-Brualla, Snavely, and Funkhouser]{wang2021ibrnet}
Qianqian Wang, Zhicheng Wang, Kyle Genova, Pratul~P Srinivasan, Howard Zhou, Jonathan~T Barron, Ricardo Martin-Brualla, Noah Snavely, and Thomas Funkhouser.
\newblock Ibrnet: Learning multi-view image-based rendering.
\newblock In \emph{CVPR}, 2021.

\bibitem[Wang et~al.(2019)Wang, Sun, Liu, Sarma, Bronstein, and Solomon]{wang2019dynamic}
Yue Wang, Yongbin Sun, Ziwei Liu, Sanjay~E Sarma, Michael~M Bronstein, and Justin~M Solomon.
\newblock Dynamic graph cnn for learning on point clouds.
\newblock In \emph{ACM Transactions on Graphics}, 2019.

\bibitem[Wang et~al.(2022)Wang, Guizilini, Zhang, Wang, Zhao, and Solomon]{wang2022detr3d}
Yue Wang, Vitor~Campagnolo Guizilini, Tianyuan Zhang, Yilun Wang, Hang Zhao, and Justin Solomon.
\newblock Detr3d: 3d object detection from multi-view images via 3d-to-2d queries.
\newblock In \emph{CoRL}, 2022.

\bibitem[Wu et~al.(2021)Wu, Chen, Fuh, and Liu]{wu2021learning}
Jhih-Ciang Wu, Ding-Jie Chen, Chiou-Shann Fuh, and Tyng-Luh Liu.
\newblock Learning unsupervised metaformer for anomaly detection.
\newblock In \emph{ICCV}, 2021.

\bibitem[Xie et~al.(2022)Xie, Yu, Zhou, Philion, Anandkumar, Fidler, Luo, and Alvarez]{xie2022m}
Enze Xie, Zhiding Yu, Daquan Zhou, Jonah Philion, Anima Anandkumar, Sanja Fidler, Ping Luo, and Jose~M Alvarez.
\newblock M$^2${BEV}: Multi-camera joint 3d detection and segmentation with unified birds-eye view representation.
\newblock \emph{arXiv preprint arXiv:2204.05088}, 2022.

\bibitem[Xie et~al.(2023)Xie, Wang, Liu, Zheng, and Jin]{xie2023pushing}
Guoyang Xie, Jinbao Wang, Jiaqi Liu, Feng Zheng, and Yaochu Jin.
\newblock Pushing the limits of fewshot anomaly detection in industry vision: Graphcore.
\newblock \emph{arXiv preprint arXiv:2301.12082}, 2023.

\bibitem[Yin et~al.(2021)Yin, Zhou, and Krahenbuhl]{yin2021center}
Tianwei Yin, Xingyi Zhou, and Philipp Krahenbuhl.
\newblock Center-based 3d object detection and tracking.
\newblock In \emph{CVPR}, 2021.

\bibitem[Yu et~al.(2021)Yu, Ye, Tancik, and Kanazawa]{yu2021pixelnerf}
Alex Yu, Vickie Ye, Matthew Tancik, and Angjoo Kanazawa.
\newblock pixelnerf: Neural radiance fields from one or few images.
\newblock In \emph{CVPR}, 2021.

\bibitem[Yuan et~al.(2023)Yuan, Gu, Li, Dong, and Zhu]{yuan20233d}
Weihao Yuan, Xiaodong Gu, Heng Li, Zilong Dong, and Siyu Zhu.
\newblock 3d former: Monocular scene reconstruction with 3d sdf transformers.
\newblock \emph{arXiv preprint arXiv:2301.13510}, 2023.

\bibitem[Zaal et~al.(2021)Zaal, Tuytel, Cilliers, Cock, Mischok, Majboroda, Savva, and Burger]{zaal2021polyhaven}
Greg Zaal, Rob Tuytel, Rico Cilliers, James~Ray Cock, Andreas Mischok, Sergej Majboroda, Dimitrios Savva, and Jurita Burger.
\newblock Polyhaven: a curated public asset library for visual effects artists and game designers, 2021.

\bibitem[Zhang et~al.(2024)Zhang, Herrmann, Hur, Polania~Cabrera, Jampani, Sun, and Yang]{zhang2024tale}
Junyi Zhang, Charles Herrmann, Junhwa Hur, Luisa Polania~Cabrera, Varun Jampani, Deqing Sun, and Ming-Hsuan Yang.
\newblock A tale of two features: Stable diffusion complements dino for zero-shot semantic correspondence.
\newblock In \emph{NeurIPS}, 2024.

\bibitem[Zhou et~al.(2024)Zhou, Li, Jiang, Wang, Zhou, Zhang, and Zhao]{zhou2024pad}
Qiang Zhou, Weize Li, Lihan Jiang, Guoliang Wang, Guyue Zhou, Shanghang Zhang, and Hao Zhao.
\newblock Pad: A dataset and benchmark for pose-agnostic anomaly detection.
\newblock In \emph{NeurIPS}, 2024.

\bibitem[Zou et~al.(2022)Zou, Jeong, Pemula, Zhang, and Dabeer]{zou2022spot}
Yang Zou, Jongheon Jeong, Latha Pemula, Dongqing Zhang, and Onkar Dabeer.
\newblock Spot-the-difference self-supervised pre-training for anomaly detection and segmentation.
\newblock In \emph{ECCV}, 2022.

\end{thebibliography}
